# A scale of conceptual orality and literacy: Automatic text categorization in the tradition of 'Nähe und Distanz'

**Volker Emmrich** (JLU Gießen)
volker.emmrich@germanistik.uni-giessen.de

**Abstract**

Koch and Oesterreicher's model of 'Nähe und Distanz' ('Nähe'= immediacy, conceptual orality; 'Distanz' = distance, conceptual literacy) is constantly used in German linguistics. However, there is no statistical foundation for use in corpus linguistic analyzes, while it is increasingly moving into empirical corpus linguistics. Theoretically, it is stipulated, among other things, that written texts can be rated on a scale of conceptual orality and literacy by linguistic features. This article establishes such a scale based on PCA and combines it with automatic analysis. Two corpora of New High German serve as examples. When evaluating established features, a central finding is that features of conceptual orality and literacy must be distinguished in order to rank texts in a differentiated manner. The scale is also discussed with a view to its use in corpus compilation and as a guide for analyzes in larger corpora. With a theory-driven starting point and as a 'tailored' dimension, the approach compared to Biber's Dimension 1 is particularly suitable for these supporting, controlling tasks.

# 1. A scale of conceptual orality and literacy for text categorization

## 1.1 Introduction

Koch and Oesterreicher's model (1985, English version: 2012, Fig. 1) of 'Nähe und Distanz' ('Nähe' = immediacy, conceptual orality: a printed interview (d) or a phone call with a friend (b); 'Distanz' = distance, conceptual literacy: an article in a newspaper (j) or a speech (i)) is constantly used in German linguistics. "The scales of communicative immediacy [...] have inspired a good deal of research not only on Romance languages but also on German and, recently, on English." (Kytö, 2019: 139) Theoretically, it is stipulated, among other things, that written texts (graphic) can be ranked on a scale of conceptual orality and literacy by linguistic features. This article establishes such a scale (conceptual-orality-literacy-scale = COL-scale) for practical use based on principal component analysis (PCA) and combines it with automatic analysis. Compared to Biber's multidimensional analysis (1988), there is a partially different area of application – not in the research of registers themselves, but in corpus compilation, orientation in large corpora and as an explanatory variable. With a theory-driven starting point and as an exaggerated, 'tailored' dimension, the approach is suitable for controlling for other factors in the study of a linguistic phenomenon. Automatic analysis is important for several reasons: In theory, only the correlation of several features, all of which have to be analyzed, justifies an abstract concept such as 'Nähe und Distanz'. In practice, during corpus compilation there are no resources for manual data analysis, but conceptual orality and literacy – among other dimensions of the texts – can play a leading role in corpus compilation. And finally, the actual research interest takes priority in human working time over the analysis of a COL-value as a resource of orientation or as an explanatory variable. But if the concept of 'Nähe und Distanz' continues to find its way into corpus linguistics, and it is entitled to do so given its research tradition (cf. Feilke & Hennig, 2016), it cannot remain theoretical.

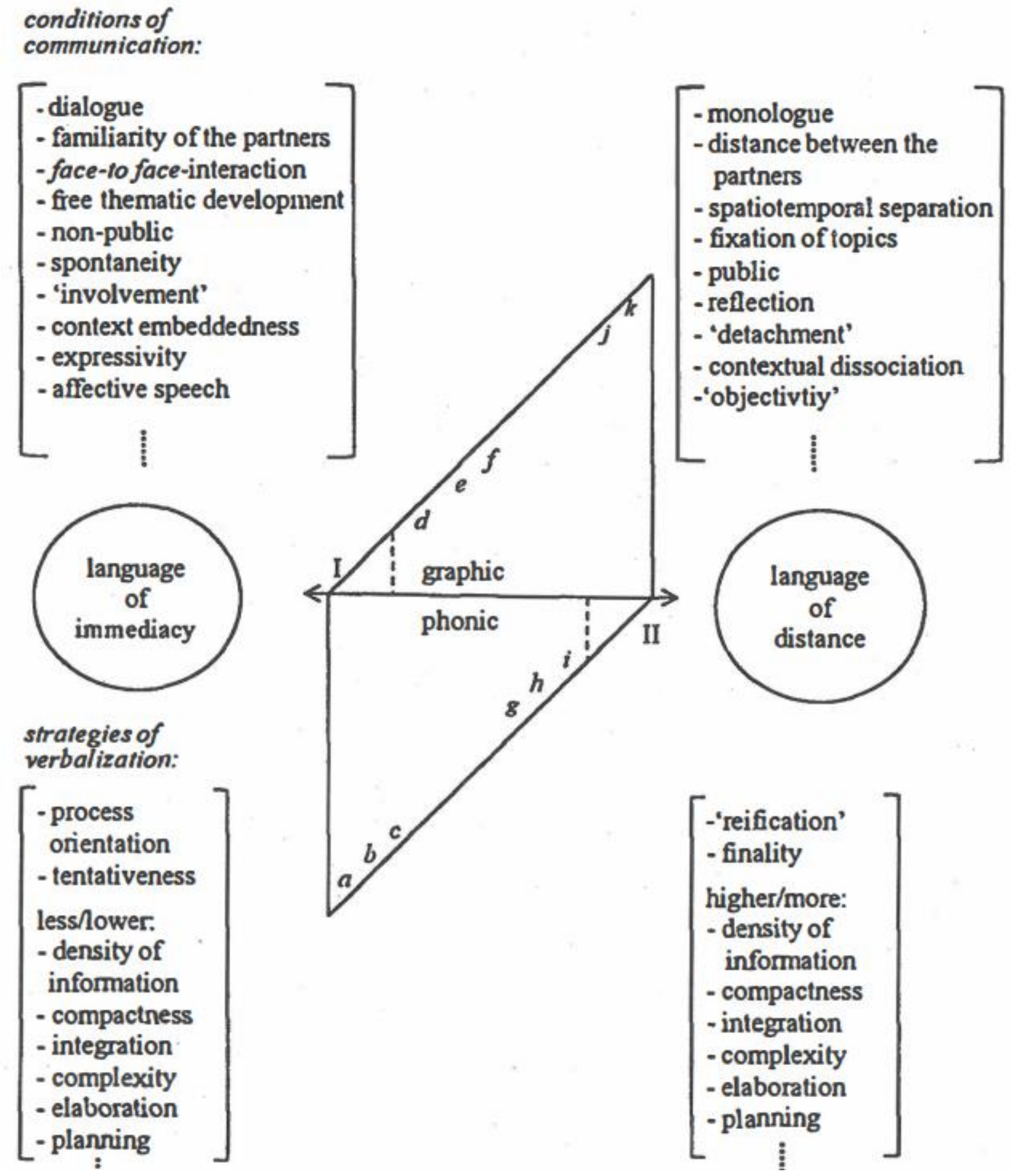

Fig. 1: The model of immediacy and distance by Koch and Oesterreicher (1985/2012)

The requirement for the correlation of features fits with a central idea of multidimensional analysis: "[It can] be misleading to concentrate on specific, isolated [linguistic] markers without taking into account systematic variations which involve the co-occurrence of sets of markers" (Brown & Fraser, 1979:38f.; see Biber, 2014), and more selectively than fundamentally, the article allows references to research on register and multidimensional analysis or, more specifically, questions of variation under the influence of spoken and written language (e.g. Biber, 1986, 1988; 2001, Biber & Conrad, 2009; Sardinha & Pinto, 2014; Seoane & Biber, 2021; more practical/methodical: Laippala et al., 2019; Biber & Egbert & Keller, 2020; Biber et al., 2021). But I would like to focus on the theory-based model of Koch and Oesterreicher. This is for the following reason: The title of a recent article by Schaeffer (2021) stands for itself: "Communicative Distance: The (Non-)Reception of Koch and Oesterreicher in English-Speaking Linguistics". With reference to Raible (2019), who compares Koch and Oesterreicher (1985) and Biber (1986), she singles out that the starting point of Koch and Oesterreicher is theory-driven, while Biber's is data-driven and she concludes (2021: 15) with reference to Raible that the genres of Biber are 'real genres' in his corpora whereas those of Koch and Oesterreicher are 'essentially theoretical', so that they "must necessarily refrain from discussing concrete linguistic features" (Raible 2019: 166).[1] Now, with the rise of corpus linguistics, the reception of Koch and Oesterreicher became by no means data-driven, but – also as a result of the approach of Ágel and Hennig (2006), who elaborate the theory of Koch and Oesterreicher most extensively for practical manual analysis, – it is increasingly being applied practically and increasingly operationalized. However, while Ágel and Hennig present a group of features that were derived from the (external) conditions of communication, empirical analyzes of individual features between theoretical poles of conceptual orality and literacy (as a consequence of a single analyzed feature) are problematic in the context of corpus linguistics.

Perhaps this situation can be described as an established tradition of the application of the theory in the context of analysis and argumentation. So this theoretical tradition must be taken up, but requires a statistical foundation. Because the (counter-)part of the problem presented here is that Biber's data-driven approach in the sense of multidimensional analysis is less used in articles that use/prefer the concept of Koch and Oesterreicher. This is confirmed by a look at Google Scholar: 3035 texts cite Koch and Oesterreicher (1985) and only 172 of them also mention Biber. However, while the research tradition of Biber (1986) emerged from technical development, that of Koch and Oesterreicher requires adaptation in order to make it easier to connect to corpus linguistics.[2] And in the course of this adjustment, both approaches appear quite similar. As Raible (2019: 172) concludes with respect to the scale of 'Nähe und Distanz' and Biber's dimension one: "Totally different approaches have led to comparable ordered scales of text genres." In my opinion the definition of external and internal features by Koch and Oesterreicher contrasts with the fact that Biber's factors ultimately have to be interpreted, and he describes dimension 1 as 'interactive vs. edited text' (1986), 'involved vs. informational production' (1988) or 'oral vs. literate discourse' (Biber & Conrad, 2009) (cf. Raible, 2019). While Koch and Oesterreicher set the position of texts on a scale, Biber relies on the original classification of the texts in the corpus (Jucker, 1989). Both Biber and Ágel and Hennig take up previous research to relate grammatical features to their possible functions. At some point, interpretation must always go beyond the analysis of the surface of the text.

[1] With their more extensive comparison of the two approaches and research traditions, both articles (Raible, 2019 and Schaeffer, 2021) offer numerous additional starting points that cannot be addressed here.

[2] As Mair (2018: 16) fundamentally states for corpus linguistics, it should also be oriented towards research interests before they adapt to the technical possibilities.

The aim here is to give the theory-driven model an empirical foundation. And that is why it is crucial to adopt Koch and Oesterreicher's established connection of internal and external features: the linguistic means must reflect the conditions of communication. Accordingly, I take the features derived by Ágel and Hennig (2006) as a starting point, since they are part of that tradion. But the model must also be combined with the ideas and methods of the data-driven approaches following Biber (1986). I ultimately give priority to Sigley (1997) and PCA over Biber and CFA when it comes to implementation. But that is not crucial here and in principle I mainly see similarities in the central analysis steps, which is also addressed in Section 1.2.

Recent articles from the field of German linguistics that attempt to implement 'Nähe und Distanz', as here, do not implement the concept of the scale (Broll & Schneider, 2023) or forego the advantages of PCA or similar calculation methods when deriving the scale (Ortmann & Dipper, 2024). As a binary classifier, Broll and Schneider's approach is not only problematic in practical application, but also a deviation from the theoretical tradition of 'Nähe und Distanz' and here the question arises as to why Biber's empirical approach is not used in the first place. Both articles model the features for prediction very simply and close to the surface of the text. Beside obvious advantages, this can also mean that further orientation via the index value only leads (back) to these features. The 9 features by Ortmann and Dipper (2024) include the mean word length, the number of demonstrative pronouns in addition to the number of *die* as demonstrative pronoun, response particles and the interjections *oh, oh, uff, ha*. In my opinion, the added value of such an index value in the form of a single number based on a few automatically analyzed features is that this analysis can be used to draw conclusions about other features that may be more difficult to analyze (by machine). But with simpler, more abstract features, further orientation is even more dependent on the existence of a correlation of these features. In the following, I take the view that more features are better as long as the dimension of 'Nähe und Distanz' remains stable, multicollinearity is avoided, and the scale is differentiated/the texts are ranked in a differentiated manner. With a view to the further analyzes here, the features of Ortmann and Dipper (2024) raise the question of whether, in addition to the area of conceptual orality, the area of conceptual literacy is also adequately covered and how the internal features represent the external features/conditions of communication. This again raises the question of the distinction from Biber's approach.

Anyway, these recent articles indicate that such an index value is relevant for corpus linguistic studies. And while Biber's approach (1988) or this entire research tradition is not taken up at all in these articles, I see a practical implementation of 'Nähe und Distanz' as an addition in the sense that with an exaggerated, tailored 'dimension' characteristics of texts can be controlled in the analysis. In my opinion, this approach can be used in the process of corpus compilation, for text selection and initial orientation before the actual analyzes (simple scalar value and the raw features behind it) and ultimately, in a first approximation, it can serve as an explanatory variable, too. Such additional comparative values are particularly important for historical corpora if a balanced, representative compilation cannot be offered. At the beginning, Kytö's article 'Register in historical linguistics' (2019) was mentioned with reference to the relevance of Koch and Oesterreicher's model (1985). As an exmaple she mentions Rodríguez-Puente (2019) study on the variation and development of phrasal verbs in English as "a recent large-scale empirical investigation of linguistic change making fullfledged use of the register parameter." (Kytö, 2019: 148) Kytö (2019: 138) also focuses on the introduction of multidimensional analysis as "an important advance in historical register studies" and on the following pages she presents the role of registers in historical linguistics in detail. She points out that "it may not be very meaningful to attempt to represent historical language use in its entirety", that "register-specific corpora has proved a promising avenue" (Kytö, 2019: 140) or that "[s]tudies based on several registers have offered convincing evidence of trends of development across the history of English.” (Kytö, 2019: 142) Here we can move on to a consideration by Imo (2016). Imo (2016: 179 f.) emphasizes the advantages of the Model of ‘Nähe und Distanz’ with reference to the project work of Ágel and Hennig (cf. 2007b) as preparatory work for a grammar of New High German. With such a grammar, according to Imo, the fiction of a uniform grammar for a certain historical stage of the language would be replaced by historical data that could be classified based on their level of conceptual orality and literacy. He concludes, for example, that the model shows its strength precisely when a corpus is created, which is thus expanded to include a lot of additional information about this text. Here I see the expansion of the corpus, for the reasons explained, first of all in automatic analyses. Koch and Oesterreicher themselves later emphasize (2007: 367 f.) that the conceptual gradations between texts should be interesting for corpus linguistic analysis – especially in diachronic research. And they underline the obvious, but ultimately by no means unproblematic, with references to literature and an example: ‘that information about the orality of past eras can only be obtained through graphically fixed texts that tend to be oriented towards language of immediacy’. They refer, for example, to Tagliavini's (1998) "Quellen zur Kenntnis des sogenannten Vulgärlateins" (Engl.: ‘Sources of knowledge about so-called Vulgar Latin’).

In the practical part, features from Ágel and Hennig (2006) become the basis for further analysis. This analysis (Binary Logistic Regression, PCA, K-mean-cluster-analysis)[3] will show that several aspects should be taken into account when developing this scale: Quality as a binary predictor, correlation of features, the ‘determination’ of PC1 on conceptual orality and literacy, the differentiation of features of conceptual orality and literacy in the course of a differentiated description of all texts. Binary classification is carried out using k-mean-cluster-analysis and it is consistently unproblematic. Feature selection is supported by a graphical review of the first two principal components. After selecting the features, 24 texts[4] (about 300,000 tokens) from the GiesKaNe corpus[5] – GIESsen KAssel NEuhochdeutsch – (Emmrich & Hennig, 2022), a syntactically deeply annotated treebank of New High German, are used to form a scale. Their relative position to each other defines the scale, which in Koch and Oesterreicher’s model corresponds to the scale that lies between the text types printed interview (Fig. 1, d) and administrative regulation (Fig. 1, k). Other texts are ranked relative to this distribution of texts or text values. The scale is stable, but – in accordance with the theory of Koch and Oesterreicher – variable in that it can change slightly with each new text. ‘New’ texts are, for example, those from the Deutsches Textarchiv (DTA[6]), which fit those in the GiesKaNe corpus, when it comes to the historical periods. The DTA is the most extensive corpus for New High German (see Fn. 6), but it is only annotated with automatic PoS-tagging and its text categories are not always consistent and precise. In this respect, the COL-value proposed here can also be seen as further assistance in using this extensive and valuable resource. However, due to the (currently) small set of texts from the KAJUK corpus[7] – KAsseler JuntionsKorpus – (Ágel & Hennig, 2007b) and the GiesKaNe corpus, the COL-scale is derived from a synchronic perspective of New High German, which means that diachronic variation is also part of the variation of the scalar values. A more precise differentiation would require more extensive data sets.

---

[3] For the analysis and calculation methods used in the present analysis, see Moisl (2015: 159 ff.) and Gries (2021: 319 ff.).
[4] https://gieskane.com/korpustexte/
[5] https://annis.germanistik.uni-giessen.de/
[6] https://www.deutschestextarchiv.de/; DTA-Kernkorpus (‘core corpus‘): 1,468 works, 150 million tokens; DTA-Erweiterungen (‘extensions‘): 5,104 works, 220 million tokens; https://www.deutschestextarchiv.de/doku/ueberblick.
[7] https://www.uni-giessen.de/de/fbz/fb05/germanistik/absprache/sprachtheorie/kajuk/kajuk-korpus; https://www.laudatio-repository.org/browse/corpus/nCQsCnMB7CArCQ9CDmun/corpora

In the next step, the entire DTA is parsed syntactically with the GiesKaNe dependency parser and ranked according to the COL-scale. Here the scale is critically examined again. The same approach is taken in parallel with ‘prefabricated’ fast-text-vectors (Joulin et al., 2016) for, among other things, text classification, which – defined via the GiesKaNe texts in PCA – deliver well-comparable results. This not only shows that these vectors can (somehow) represent abstract concepts from theoretical linguistics. When applied to the DTA, this analysis also becomes practically relevant to the discussion in section 2.3 and can therefore also be seen as a tool for linguistic analysis. A comparable analysis is carried out by Crossley and Louwerse (2007), who use bigrams and thus also present a less theory-driven approach to text classification:

> “The advantage of a bigram approach is that it does not assume syntactic information, but rather lexical information by taking frequent collocations in different corpora. At the same time, however, the bigrams isolated here are not only based on lexical differences, but also on more latent, syntactic and discourse features.” (Crossley & Louwerse, 2007: 474)

In addition, with a view to the practical work on the GiesKaNe corpus and in comparison to the DTA, a smaller sub-study asks how well the shorter text sections in GiesKaNe represent the entire texts, some of which are included in the DTA corpus. This is relevant because the respective metadata will be fully used to explain the variation in the analyzes of these sections. Finally, as a consequence of the orientation towards Koch and Oesterreicher’s concept, the COL-values are compared with external conditions of communication/the external features of the text (e.g. monological vs. dialogical communication, familiarity vs. unfamiliarity of speaker and listener, face-to-face-interaction vs. spatial and temporal separation, see Fig. 1), if this is possible on the basis of text types and individual text examples.

### 1.2 Previous research

Since this article aims to bring together different research traditions and their stages, the contributions and topics discussed here are different but complementary. Ágel and Hennig (2006), who elaborate the theory of Koch and Oesterreicher most extensively for practical manual analysis, discuss over 40 features[8] (Ágel & Hennig, 2007a:189ff.) and they rate 8 texts in the KAJUK corpus. As a consequence of a selection of features to operationalize Nähe ‘versus’ Distanz, these texts tend to group almost exclusively at the ends of the scale. The 8 texts (6 conceptually oral and 2 conceptually written texts) show the following values: 43.4, 39.3, 39, 38.6, 35.3, 29 and 4.1, 2.6. While Ágel and Hennig seem to have a single, absolute scale in mind in terms of the addition of feature scores, I see a flexible scale and a relative positioning in the theory of Koch and Oesterreicher. Accordingly, Ágel and Hennig’s maximum is 43.4 and the minimum is 2.6. Except for the text Zimmer with 29, only the ends of the scale are occupied. You can think of the idea of relativity here in such a way that, on the one hand, this scale on the basis of the texts from the GiesKaNe corpus only fits corresponding texts from New High German (more on that later) and, on the other hand, one can ask the question what would happen if one were to throw a hypothetical/theoretical text type/form of communication ‘whatsapp/facebook-messenger message’ into Koch and Oesterreicher’s theoretical scale of medial written texts between a printed interview and an administrative regulation. Firstly, this would probably be the new pole of conceptual orality, and secondly, the other texts would move together if the representation of the theoretical scale (as a line, for example) remained unchanged. In my opinion, this relativity corresponds to the model of Koch and Oesterreicher and the theoretically modeled scale with its text types and PCA will implement this property well.

[8] If subcategories are included, the number can be described as even more extensive.

And this indicates another problem, which becomes clear in the analyzes in Section 2. The approach of Ágel and Hennig (2006) only aims to elaborate the area of conceptual orality for practical application. As a consequence, the part of the scale that concerns the conceptually written/literal texts is not used in a differentiated manner if, for example, interjections, personal pronouns for speaker and listener, (deictic) time references generally occur very rarely. This means that these values are often close to zero. This should also apply to 3 out of 9 features in Ortmann and Dipper (demonstrative pronouns with the short word form *die*, response particles and interjections) and can be seen here in Fig. 12 in the column interjections. In this article, a distinction is therefore made between features of conceptual orality and features of conceptual literacy. Those of conceptual orality by Ágel and Hennig (2006) are tested for significance and relevance and for the possibility of automatic analysis. Those of conceptual literacy are added as a first suggestion. The selection is made selectively and without too extensive theoretical specifications , but justifiable. However, it would have to be sensibly substantiated by an analysis such as that by Ágel and Hennig (2006) for feature of conceptual orality. It turns out that some of the features of Ágel and Hennig (2006) have no effect on the distinction between conceptual orality and conceptual literacy; in the case of parenthesis, there is even a negative correlation with other features of conceptual orality. Their calculation method must also be questioned because all features are just counted and summed up. Here, however, the scale is derived directly from the first principal component of PCA: All features are included in a text vector, scaled and normalized. All features are included in a text vector and standardized using StandardScaler()[9], ensuring zero mean and unit variance before PCA. This approach is less linked to Biber's research (1985, 1986, 1988) in the sense of multidimensional analysis via factor analysis (CFA), but is similar to Sigley (1997), who derives a crude formality index in the application of PCA. Advantages of this approach are also highlighted in the detailed discussion of the two recent articles from German linguistics.

There are also approaches to automatic implementation from the research tradition of multidimensional analysis: For example, automation of factor analysis according to Biber (1988) by Nini (2019) or text classification by Repo and Hashimoto and Laippala (2023). But I will focus on the ones related to concept of 'Nähe und Distanz'.

Broll and Schneider (2023) train a binary classifier based on features that are closely linked to the text surface supported by a simple PoS-tagging and they focus on the evaluation of these features. So they cannot build on the work of Ágel and Hennig (2006). Broll and Schneider's poles of conceptual orality and conceptual literacy are represented by transcripts of spoken language and newspaper texts. As Sigley (1997: 219 f.) shows, there are differences between medial orality and medial literacy. Therefore, in my opinion, the question of the extent to which conceptual orality can be derived from medial orality should be addressed in more detail here. Although it should not be practically relevant to Broll and Schneider's binary classification, it would be if a scalar classification were used instead of a simple binary. The further analyzes here, such as the distinction between features of conceptual orality and literacy with the aim of achieving a balanced relationship, show the strong effect that features have that do not (cannot) appear in certain types of text. As already mentioned with the relativity of the scale, the question arises as to what should be ranked. In this article, for example, New High German texts from the DTA with reference to the GiesKaNe treebank. So it is all about texts in the medium of writing. When creating a scale between everyday conversations and newspapers, these problems would have to become even more apparent.

[9] https://scikit-learn.org/stable/modules/generated/sklearn.preprocessing.StandardScaler.html

Due to  Broll and Schneider binary classification, other text types (between the poles) can only be arranged based on the proportion of individual texts assigned to the two groups. My approach focuses more on the theory of Koch and Oesterreicher (1985) and the COL-scale via PCA with an additional binary classification of the texts. The latter is unproblematic in implementation (cf. Ortmann & Dipper, 2024: 30), but in practice it is too imprecise. With a weaker orientation towards Koch and Oesterreicher's model, a distinction from Biber (1986) is ultimately missing.

Ortmann and Dipper (2024, 2020, 2019) also rely on surface-level-operationalization and they have to discard the approach of Ágel and Hennig (2006) as only possible to implement manually. Unlike Broll and Schneider (2023), they convert the analysis into a scalar value and apply it to the 8 texts of the KAJUK corpus. As here, a set of 9 features is selected based on the text's binary classification by experts. But then the normalized values of their 9 features are multiplied by their information gain value and the results are added together. Some values are included as negative numbers (Ortmann & Dipper, 2024: 32). While the assumption that higher average word length indicates lower conceptual orality is initially plausible, one can ask whether it would be better to statistically prove a sign change for the operationalized information gain values, especially since previously in a linear Regression model for conceptual oral texts alone, only 3 features show up with a significant contribution. However, this analysis then appears to have no effect on the subsequently calculated score. In my opinion, the fact that they do not rely solely on an efficient model or the best predictors is not a critical point. A supplementary goal for the COL-value will be a differentiated use of both segments of the scale (conceptual orality and conceptual literacy) – preferably using the same feature of a text. However, the interaction of the individual features remains the central aspect in which the advantages of PCA become apparent. As mentioned, the features of Ortmann and Dipper (2024) appear more practical than related to the external conditions of communication and 3 features could occur especially in texts of conceptual orality. The comparison of the COL-values with those of Ágel and Hennig (2006) and Ortmann and Dipper (2024) at the end of section 2.2 will show comparability of the values. But it will also become clear that in both previous approaches, primarily the edges of the scale are used. While in Ágel and Hennig (2006) theoretically justified but unevaluated features have to be laboriously analyzed manually, the features according to Ortmann and Dipper (2024) can ultimately be emphasized in terms of practicality.

In this article, the PCA corresponds to the calculation steps described in Ortmann and Dipper (2024), whereby PC1 forms the scale and integrates all features (in reverse with linear regression, for example) in a highly significant and relevant manner. The ratio between (1.) the texts that were set to form the scale in accordance with the theory, (2.) the evaluated features and (3.) the COL-values is optimized. Sigley (1997: 212) highlights the advantage of PCA when he compares PCA with CFA and Biber's approach (1988):

> "Biber's method (summing standardised counts with loading magnitudes greater than 0.35 on the factor, excluding counts with higher on other factors) is arbitrary, so defensible in terms of maximising factor interpretability. By contrast, PCA uniquely defines its summary factors as a weighted sum of all standardised counts (where the weights are the PCA 'loadings')."

Sigley (1997: 215 ff.) himself selects 29 features for his crude formality index via PCA and then looks at the loadings. He expects "most counts to load highly on PC1, providing an index for the shared function", while "subsequent principal components […] represent variation 'left over' from the required index". PCA in the present article only follows a selection of features as binary predictors, which, in my understanding, should define PC1 even more clearly to the desired distinction between conceptual orality and conceptual literacy. Sigley (1997: 230) also emphasizes in conclusion:

> "The number of counts could also be reduced, for example, by dropping all counts with a loading of less than 0.5 on PC1 […]. Such changes will consolidate the index as a multivariate but unidimensional score, losing any advantages that PC2 and subsequent principal components may offer for overall comparisons."

Here, the loadings of PC1 represent the numerical ratios of the features in Fig. 12. Sigley's PC1 as a formality index explains 41% of the variation in 29 features. Here PC1 explains 52% of the variation of ultimately 9 features in the intended COL-dimension.

In comparison to Biber (1986, 1988), what stands out is that, in the spirit of the data-driven approach, he works with a less restricted set of over 41 or 67 features and aims to describe several factors:

> "These 67 features represent several form-function pairings; features from the same grammatical category can have different functions, and features from different grammatical categories can have a shared function. As such, these features provide a solid basis for determining the underlying functional dimensions in English." (Biber, 1988: 72)

On the one hand, Ágel & Hennig also use well over 40 such features, despite their focus on conceptual orality, and on the other hand, the assignment to specific functional areas is not unproblematic with 9 features, as here (see section 2). A more general comparison between Biber's features and those used here in Section 2.2 will present the present approach to 'Nähe und Distanz' as an exaggerated, tailored dimension, as indicated by a strong PC1 versus a weak PC2, while Biber's (1988) Dimension 1 features applied to the GiesKaNe texts, explain less variation and PC1 and PC2 differ noticeably less.

Here, as in Biber (1988: 104ff. and 223ff.) potential functions are described too much in isolation. Association patterns (as in Biber, 1996) must move further into the center of the analysis. The consideration of a large number of multifunctional grammatical features requires a stronger analysis from the perspective of concrete linguistic interaction rather than an isolated, additive one. In this sense, the present analysis should also be supplemented by one of larger lexico-grammatical patterns. When it comes to the features used, there is agreement on the 1st and 2nd person pronouns, passive voice, temporal adverbials and information density. And these also hit the core of Biber's first dimension, which he describes via 'two separate communicative parameters':

> "(1) the primary purpose of the writer/speaker: informational versus interactive, affective, and involved; and (2) the production circumstances: those circumstances characterized by careful editing possibilities, enabling precision in lexical choice and an integrated textual structure, versus circumstances dictated by real-time constraints, resulting in generalized lexical choice and a generally fragmented presentation of information." (Biber, 1988: 107)

And as we will see, they are also, unsurprisingly, central to Koch and Oesterreicher (1985) and Ágel and Hennig (2006).

# 2. The derivation of a practically applicable scale of conceptual orality and literacy

## 2.1 The starting point:Evaluation of the features of conceptual orality by Ágel and Hennig (2006)

The starting point is a data-driven approach and a vector representation of the GiesKaNe texts. For this purpose, fast text vectors (Joulin et al., 2016) from SpaCy (de_core_news_lg[10]) tailored to the task of efficient text classification are used. These make it possible to represent words, sentences and ultimately texts through further computation as 300-dimensional vectors, whereby – from a linguistic perspective with reference to Wittgenstein and Firth and greatly simplified – words are represented more abstractly through their contextual conditions, detached from grammatical theory, so that they can be used more diversely and flexibly. Document vectors are generated by averaging the word embeddings of each text (Mean Pooling). PCA and k-mean cluster analysis based on these vectors are used to graphically place these texts in two dimensions and to divide texts into two groups. Figure 2 shows the results.

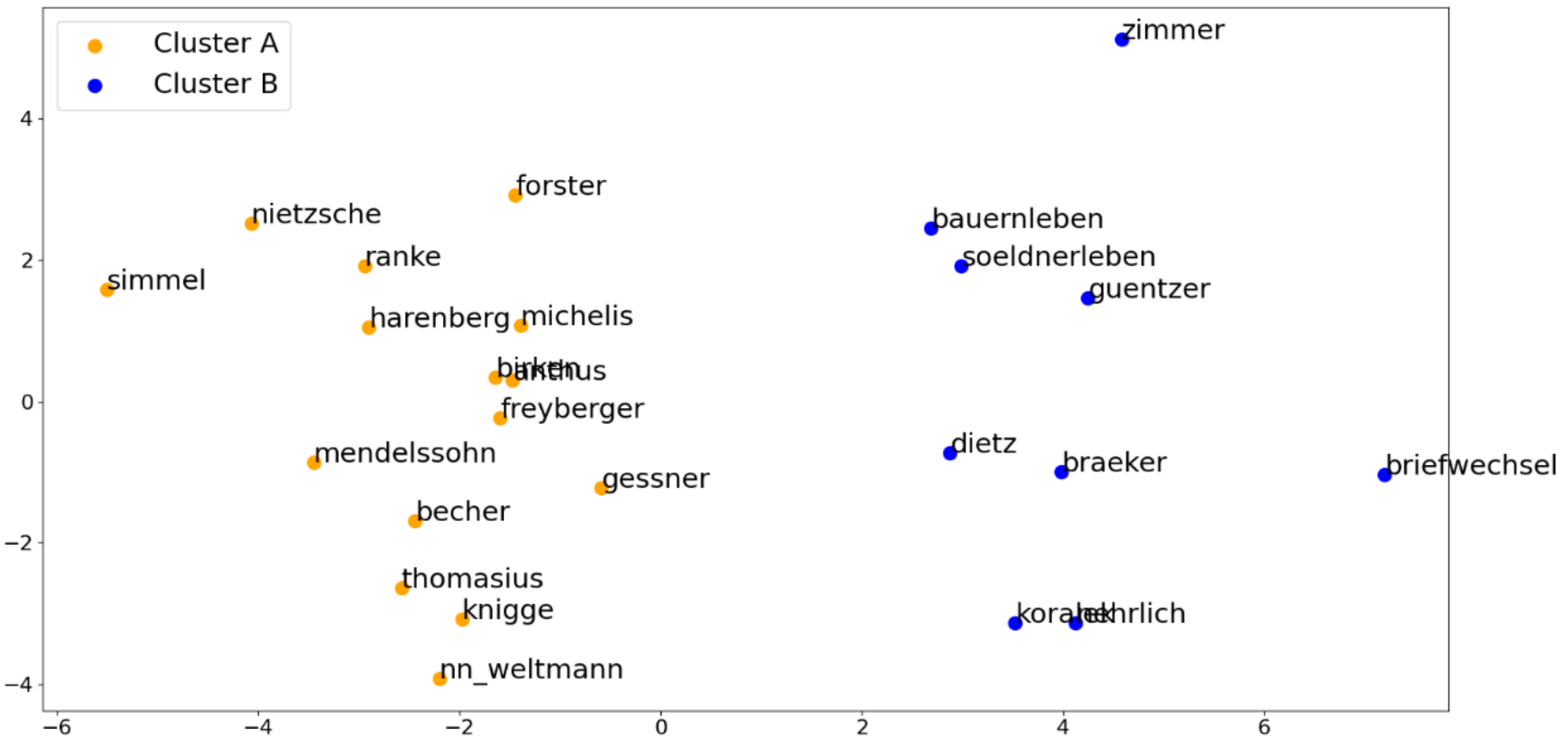

Fig. 2: 2D representation via PCA and cluster analysis of fast-text vectors for 24 GiesKaNe-texts

The binary assignment to one of the two clusters in yellow and blue divides the texts between conceptual orality and literacy exactly according to the human analysis by Ágel and Hennig (2006) as in the KAJUK corpus (Ágel & Hennig, 2007b) and in the GiesKaNe corpus (Emmrich & Hennig, 2022). This analysis is compared with that based on 9 selected features in Section 2.2. This data-driven analysis using fast-text vectors, so to speak, empirically neutralizes and consolidates the binary classification by experts in the GiesKaNe corpus so that, for example, when analyzing the features according to Ágel and Hennig's approach, the interaction between features and the initial classification does not need to be given special consideration. Fast-text vectors will be taken up again in Section 2.3 and will be used practically in discussion.

Theory-driven, 21 of the features by Ágel and Hennig (2006), as listed in Fig. 4, are the practical starting point for implementing the COL-scale. As mentioned, the features justified by Ágel and Hennig are more numerous. These are the features that can be evaluated in terms of significance, relevance and automatic analysis using the GiesKaNe treebank. Based on the 21-dimensional vectors, a new PCA and a cluster analysis are made (Fig. 3). The binary text classification in GiesKaNe is indicated with a preceding *N* ('Nähe' = conceptual orality) or *D*

---

[10] https://spacy.io/models/de

(‘Distanz’ = conceptual literacy). The comparison of the A cluster with the B cluster shows that these 21 features were modeled with conceptual orality in mind and that variation in the texts of conceptual literacy needs to be worked out by additional features. The small gap between Dietz and Bräker on the one hand, both autobiographies (cf. Rauzs, 2006a, 2006b), and the A cluster on the other makes it clear why binary classification is problematic. The B cluster texts scattered on the x-axis are reminiscent of the theoretical concept of the scale and the possibility of smooth transitions.

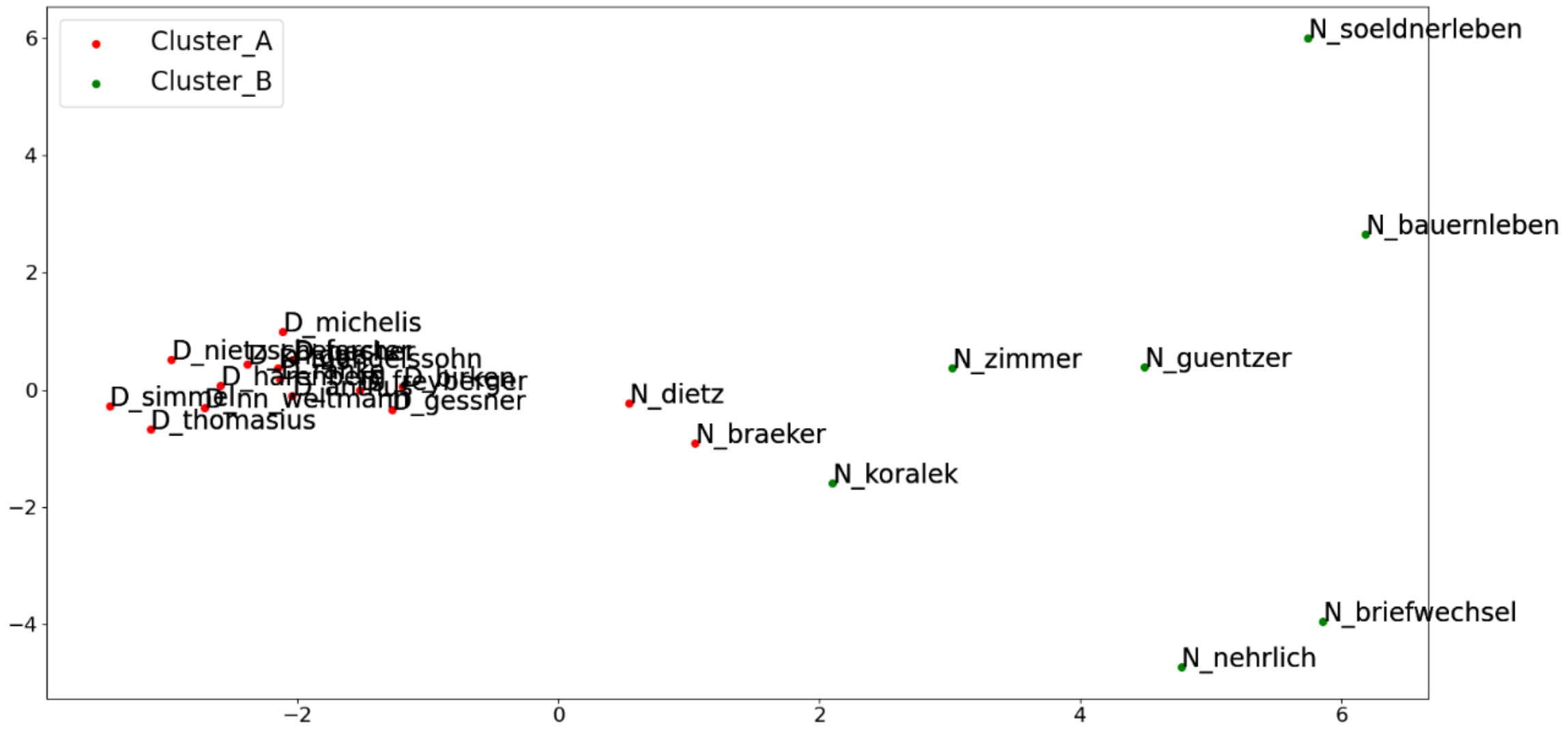

Fig. 3: 2D representation via PCA and cluster analysis of 21 features of Ágel and Hennig (2006) vectors for 24 GiesKaNe-texts

| | | LR | p ≈ | p-class | $R^2$ |
|---|---|---|---|---|---|
| 1 | vocative | 15.5 | 8.352e-05 | *** | 0.65 |
| 2 | parenthesis | 12.6 | 0.0004 | *** | 0.56 |
| 3 | structures at sentence boundary | 8.3 | 0.004 | ** | 0.4 |
| 4 | anacoluthon | 18.3 | 1.844e-05 | *** | 0.73 |
| 5 | apokoinu | 8.3 | 0.004 | ** | 0.4 |
| 6 | independent clause | 8.5 | 0.003 | ** | 0.41 |
| 7 | dependent main clause | 6.2 | 0.01 | * | 0.31 |
| 8 | interjection | 23.1 | 1.506e-06 | *** | 0.84 |
| 9 | correlative | 0.0 | 0.67 | | 0.0 |
| 10 | modal particle | 4.0 | 0.045 | * | 0.21 |
| 11 | non-verbal unit | 12.1 | 0.0005 | *** | 0.54 |
| 12 | exbraciation | 15.4 | 8.662e-05 | *** | 0.65 |
| 13 | right dislocation | 31.8 | 1.749e-08 | *** | 1.0 |
| 14 | left dislocation | 13.5 | 0.0002 | *** | 0.59 |
| 15 | verb first main clause | 9.6 | 0.002 | ** | 0.45 |
| 16 | mean sentence length | 31.8 | 1.745e-08 | *** | 1.0 |
| 17 | mean subordinate clauses per sentence | 21.5 | 3.46e-06 | *** | 0.81 |
| 18 | finit before non-finite in subordinate clause | 2.5 | 0.1 | | 0.14 |
| 19 | pronouns $1^{st}$ and $2^{nd}$ person (speaker, listener) | 31.8 | 1.745e-08 | *** | 1.0 |
| 20 | temporal adverbials | 31.8 | 1.745e-08 | *** | 1.0 |
| 21 | local adverbials | 13.4 | 0.0002 | *** | 0.58 |

Fig. 4: Binary Logistic Regression for twenty-one of the features by Ágel and Hennig (2006)

In the following Binary Logistic Regression (Fig. 4), 4 features show little potential for determining conceptual orality vs. conceptual literacy (7, 9, 10, 18), and the $R^2$ values of 4 additional features fall behind the others (3, 5, 6, 15), but that does not have to be an exclusion criterion. Here, however, they are excluded in order to ensure a stable definition of PC1 as the COL-dimension. Further features can and should be included in the vectors as long as the definition of PC1 remains stable and as long as the texts are distributed further along the scale instead of grouping at the edges. If the features are too vague or too many, PC1 is less clear, and the proportion of variation explained should decrease. After this first analysis step, 13 features remain.

The concept of the grammatical sentence is fundamental for automatic analysis using dependency parsing and further analysis based on this. In contrast to the orthographic sentence, which is delimited by a punctuation mark, the grammatical sentence represents a syntactic unit (cf. Ágel, 2017). It contains a single predicate that carries the valence information and creates the scenario, the predicate's complements and other supplements, which together create the situation. Therefore, the decrease in mean sentence length (cf. Rudnicka, 2018), as noted by Biber and Conrad (2009), could indicate a completely different phenomenon that can be explained by changing punctuation practices, but in my opinion does not necessarily have to be an expression of complexity: "Perhaps the most important change has been in the syntactic complexity typical of eighteenth-century versus modern novels. Sentence length is one measure of this difference;" (Biber & Conrad, 2009: 152) Because complexity can probably only be measured by sentence length in relation to the predicates and their scenarios and/or should at least be compared with a feature of complexity: "However, the syntactic complexity of eighteenth-century novels extends well beyond punctuation practices. One of the most important differences from modern novels involves the syntactic complexity of noun phrases." (Biber & Conrad, 2009: 154) While Rudnicka (2018) uses a variety of explanations for this trend of decreasing length, in my opinion at least a comparison with the number of predicates is missing.

Mean sentence lengths of 10.3 words[11] (sd = 1.6) for conceptually oral texts and 19.0 words (sd = 3.4) for conceptually written texts can be observed here (Fig. 5). Biber (2014, cf. Biber & Gray, 2010) describes this as

> "perhaps the most important and robust finding to emerge cross-linguistically from MD studies: Spoken registers (and 'oral' written registers) rely on clausal discourse styles, including a dense use of dependent clauses; written registers (and 'literate' spoken registers) rely on phrasal discourse styles, especially the dense use of phrasal modifiers embedded in noun phrases."

In Section 2.3, this difference is taken up again with reference to the DTA corpus.

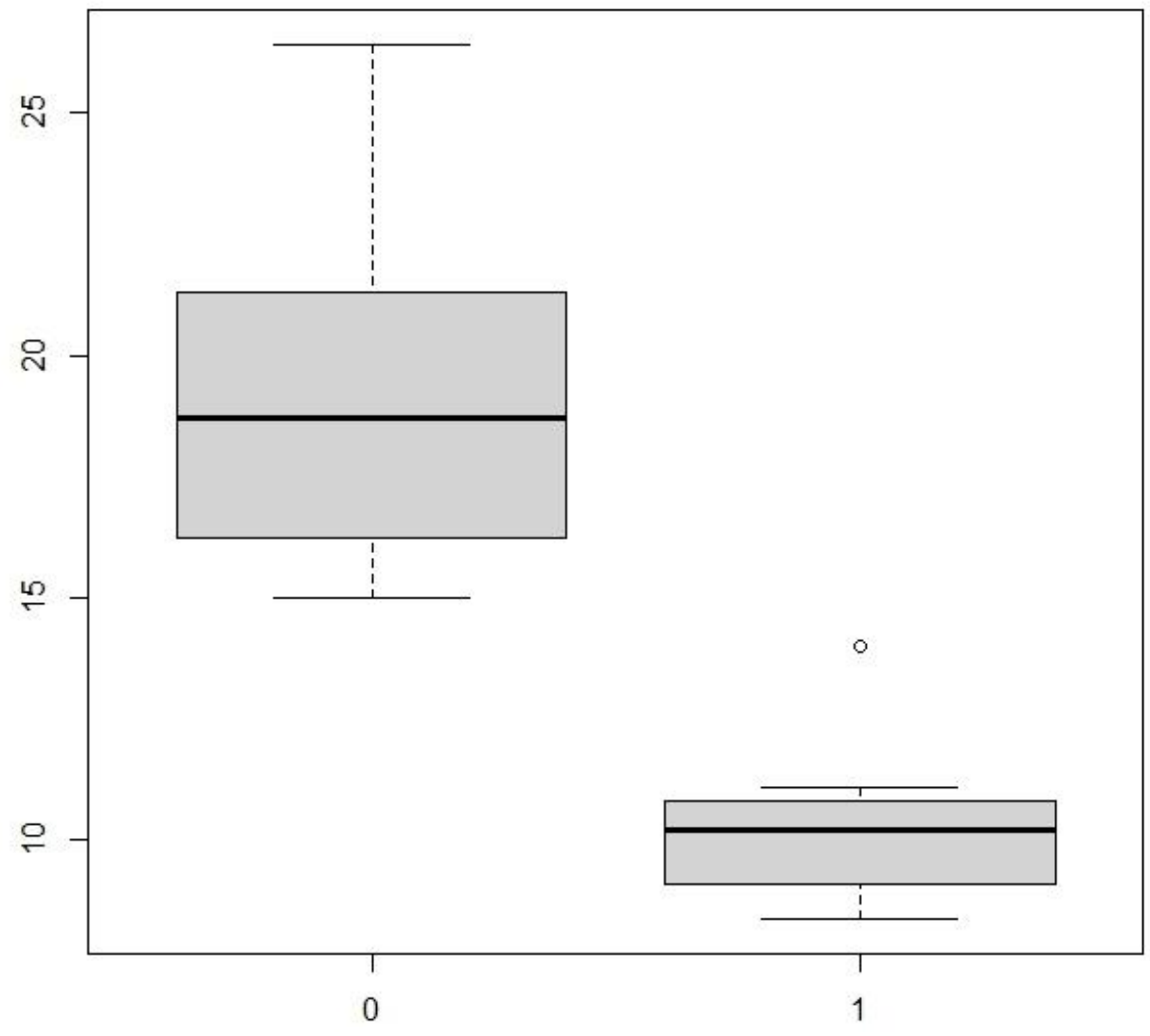

Fig. 5: Mean sentence length in GiesKaNe annotations, c. literacy = 0, c. orality = 1

[11] To normalize the frequency counts, it is advisable to exclude punctuation marks from the group of tokens due to differences in usage and frequency. Since automatic analysis is fundamental in this context, *word* is understood here as an orthographic word. In contrast, the GiesKaNe treebank uses a morphosyntactic concept of word.

Binary distinction between conceptual orality and literacy by a single independent variable is practically possible, but not central here; correlation is crucial. Only when a bundle of correlating features stabilizes to form a pattern it can be concluded that other features that are not recorded are associated with them and only this correlation can justify the abstract concept of 'Nähe und Distanz'. Ideally, internal features should of course also be linked to the external features of communication (monological vs. dialogic communication, etc.).

Figure 6 shows trends in the correlation of features. The length of the grammatical sentence is, as mentioned, a central value: the shorter the sentence, the more frequently the features of conceptual orality occur. Parenthesis can therefore be deleted because they correlate negatively with other features than sentence length and could therefore be taken up again as a positive feature of conceptual literacy. Vocatives fall behind other highly correlated features. If there is multicollinearity between the number of subordinate clauses and the grammatical sentence length, this is not relevant because both variables are ultimately combined into one variable: The first two features according to Fig. 6 are (implicitly) taken into account via the number of predicates minus the count of words (parsed, PoS-tagged) that can introduce a subordinate clause, such as subordinating conjunctions, relative pronouns and to-infinitives relative to the number of token.

| ● < p=0.001 < ● < p=0.01 | mean sent length | subordinate cl. count | no verb unit | interjection | pron. 1st, 2nd | temporal adv. | local adv. | left dislocation | r. dislocation | exbraciation | parenthesis | vocatives |
|---|---|---|---|---|---|---|---|---|---|---|---|---|
| subordinate cl. count | 0.9 | | | | | | | | | | | |
| no verb unit | -0.62 | -0.52 | | | | | | | | | | |
| interjection | -0.6 | -0.52 | ≈ 0 | | | | | | | | | |
| pron. 1st, 2nd | -0.75 | -0.56 | 0.54 | 0.63 | | | | | | | | |
| temporal adv. | -0.77 | -0.71 | 0.59 | 0.66 | 0.77 | | | | | | | |
| local adv. | -0.6 | -0.60 | ≈ 0 | 0.36 | 0.43 | 0.53 | | | | | | |
| left dislocation | -0.44 | -0.42 | ≈ 0 | 0.67 | 0.16 | 0.28 | 0.56 | | | | | |
| r. dislocation | -0.71 | -0.64 | 0.60 | 0.68 | 0.61 | 0.62 | 0.80 | 0.77 | | | | |
| exbraciation | -0.54 | -0.51 | 0.64 | 0.32 | 0.27 | 0.35 | 0.81 | 0.72 | 0.85 | | | |
| parenthesis | 0.61 | 0.4 | -0.58 | -0.52 | -0.71 | -0.46 | -0.48 | -0.37 | -0.65 | -0.48 | | |
| vocatives | -0.39 | -0.27 | 0.47 | 0.45 | 0.64 | 0.48 | ≈ 0 | ≈ 0 | 0.40 | ≈ 0 | -0.41 | |
| anacoluthon | -0.62 | -0.42 | 0.24 | 0.56 | 0.75 | 0.57 | 0.49 | ≈ 0 | 0.46 | 0.23 | -0.60 | 0.29 |

Fig. 6: Correlation (Pearson) of the features of conceptual orality, left column/header: orange = weak correlation or correlation that deviates from the pattern of conceptual orality, yellow = analysis difficult to implement, grey = possible but not included

No-verb-units, right dislocation, and anacoluthon cannot be captured well in automatic analysis. This is a group of phenomena in which the lack of orthography cannot be compensated by the effects of the verbal bracket. No-verb-units are, in their very concept, a constructive analysis. They require a determination of sentence: The same problems arise as in determining the grammatical sentence, and units without a verb are also possible features of conceptual literacy as headings. But exbraciation can be implemented here , although in some cases a distinction with right dislocation cannot/must not be drawn, and, in addition to the operationalized sentence length feature, stands as a 'representative' for the group around the topic 'verbal bracket'.[12] Local adverbials are omitted in favor of temporal ones. Of the remaining features, only exbraciation seems to require explanation (Example 1). Exbraciations are structures that occur within the sentence but at its right boundary. They are characterized by the fact that an element that can be in the middle field is realized after the right part of the verbal bracket – after the past participle (*genommen*) in example 1 (Ágel & Hennig, 2006: 389 f.; Engel, 1974; Zahn, 1991; regarding the verbal bracket, see: Smirnova, 2021).

[12] The fact that exbraciation itself drops when it comes to correlation is mainly due to a text, Söldnerleben, in which the use of this structure is so high that the values are strongly influenced. This in turn speaks in favor of using mean values for text types as in Koch and Oesterreicher. In practice, however, this seems rather unlikely

> **Example 1:**
> Deine lieben Briefe [...] habe Ich [...] in Enpfang genommen kurz vorn Abmarschieren auf Vorposten. (Briefwechsel)
> [engl.: Your lovely letters […] I have […] received shortly before setting off for the outpost.]

The length of the grammatical sentence can be viewed as the most general feature between conceptual orality and literacy and can be related to lower and greater complexity and information density. Interjections are an indicator of more or less emotionality, but can also indicate different degrees of spontaneity and planning. First and second person personal pronouns indicate dialogic structures. Spatial and temporal reference to the situation is captured by Ágel and Hennig (2006) via deictic adverbials, while it was shown here that this assumption also holds for the number of all temporal adverbials and can well distinguish conceptual orality and literacy as a binary predictor. Ágel and Hennig (2006) relate exbraciation to conceptual orality und aggregation in the pair of terms 'aggregation and integration' and to the question of whether planning and production take place at the same time or whether elements are more (semantic-pragmatic) coherent than (structural) cohesive (26 f.). The result is a justifiable selection of features, but also only a first approximation in the sense discussed. The aim must be to implement additional features in a stable manner, to exhaust the scale and at the same time maintain the dimension between conceptual orality and literacy. Since automatic analyzability was not a criterion for Ágel and Hennig (2006), one could also rework the features of conceptual orality from this perspective.

Before this set is supplemented with features of conceptual literacy, as mentioned above, the following section begins with the automatic analysis of these features and their application in PCA. It concludes by merging both groups of features and a final derivation of the COL-scale.

## 2.2 Automatic analysis of the 5 features of conceptual orality and complementary features of conceptual literacy

Automatic text classification is done by using a dependency parser and a PoS-tagger trained on the GiesKaNe corpus (UAS = 0.88, LAS = 0.77, $PoS_{Acc}$ = 0.94)[13]. However, even before any morpho-syntactic analysis of the sentence itself, determining its boundaries is critical. The length of the grammatical sentences was determined to be the clearest predictor of conceptual orality and literacy. In this context, automatic segmentation into orthographic sentences could favor longer sentences in conceptually oral texts, given the lower impact of conventions in some text types. In order to enable a comparable analysis, purely automatic sentence boundary detection is used for all texts as part of the evaluation with GiesKaNe. Two models are trained based on the annotations in GiesKaNe (24 texts, each with 80% training, 20% testing)[14]: one for orthographic sentences (F-score = 0.94) and one for grammatical sentences (F-score = 0.45). For the latter, all punctuation marks were removed from the GiesKaNe text layer. The use is sequential and, for seven texts that are used here alone for

---

for the GiesKaNe corpus with its small number of texts – especially if diachronic variation is also taken into account.

[13] SpaCy (3.0): dependency parser, https://spacy.io/api/dependencyparser, https://spacy.io/usage/facts-figures; Transformer-Model, de_dep_news_trf (3.7.2), https://spacy.io/models/de.

[14] https://spacy.io/api/sentencerecognizer

evaluation, it provides an average agreement of 77% in the assignment of the token to the sentences with the texts annotated in GiesKaNe.

The mean values are between 0.86 and 0.60 and the proportion of identically segmented grammatical sentences is between 0.59 and 0.19 (mean = 0.38) (Fig. 7). Texts like Braeuner with only 60% or 19% agreement could arise in practical use and indicate possible complications, but are by no means the standard: here one finds orthographic and grammatical sentence lengths further above the value ranges shown in Fig. 5, but also frequent, indistinct changes between New High German and Latin.

| | **mean**<br>tok ident | **standard derivation**<br>tok ident | **100 % identity**<br>grammatical sentence |
|---|---|---|---|
| goethe | 0.82 | 0.20 | 0.45 |
| becker | 0.81 | 0.22 | 0.45 |
| bosmann | 0.72 | 0.24 | 0.29 |
| braeuner | 0.60 | 0.28 | 0.19 |
| buchholz | 0.77 | 0.21 | 0.28 |
| fontane | 0.82 | 0.20 | 0.42 |
| gellert | 0.86 | 0.21 | 0.59 |

Fig. 7: Evaluation of automatic sentence boundary detection

With this automatic segmentation, conceptually oral and written texts are only slightly closer to each other: $mean_{c_orality} = 11.9$, $std_{c_orality} = 1.4$; $mean_{c_literacy} = 18.3$, $std_{c_literacy} = 2.5$; The correlation between annotation and automatic analysis is $r = 0.92$ with $p < 0.001$. In my opinion, a syntactic analysis can build well on this segmentation, as long as one carefully interprets results in the combination of certain text types with phenomena on the right boundary of the sentence, such as exbraciation. However, if the sentence boundary recognition or the sentence length is related to the feature in the text vector, it becomes clear why this feature had to be operationalized with reference to the surface through a combination of token, predicates and subordinating words (e.g. subordinating conjunctions). Even a very good analysis, which is particularly prone to errors when it comes to individual text types and centuries, would not be helpful here.

The other features are then at a higher level of accuracy: interjections ($r = 0.995$, $p < 0.001$) and speaker-listener pronouns ($r = 0.987$, $p < 0.001$) are based more on pardigmatic-categorical concepts and they are unproblematic in automatic analysis. Temporal adverbials as syntagmatic relations in the sentence can also be reproduced well in the parser analysis compared to the annotations ($r = 0.956$ $p < 0.001$). Ultimately, the exbraciation ($r = 0.978$, $p < 0.001$) also turns out to be a stable value due to a stable reconstruction of the verbal bracket. As mentioned before, special attention should be paid to the interpretation of this feature and its interactions with sentence boundary detection and individual text types. However, this does not affect the COL value itself.

As previously mentioned, operationalized sentence length proves to be a fundamental predictor for distinguishing between conceptual orality and literacy. With a view to further analysis, it therefore seems reasonable not to include it solely as a feature of conceptual orality. After PCA, 4 features of conceptual orality show a similar distribution in the 2D plot as in Fig. 3 for the PCA with 21 features and they further show a stable cluster assignment (Fig. 8a and 8b): While the texts of the B/N cluster of conceptually oral texts (green) shift somewhat internally, their edges tend to remain stable. However, the value range of the axes has been reduced significantly, so that the texts are no longer described as diverse as in Fig. 3 with 21 features. As mentioned at the beginning, the cluster of conceptually written texts has not yet been analyzed in a differentiated manner because features such as interjections, first and second person pronouns and exbraciation play little or no role here. The only apparent similarity in the values arises primarily from the fact that there are relatively uniformly low values compared to the cluster of conceptually oral texts. It appears that good binary predictors do not go hand in hand with the potential to describe the cluster of conceptually written texts in a differentiated manner in contrast to the cluster of conceptually oral texts.

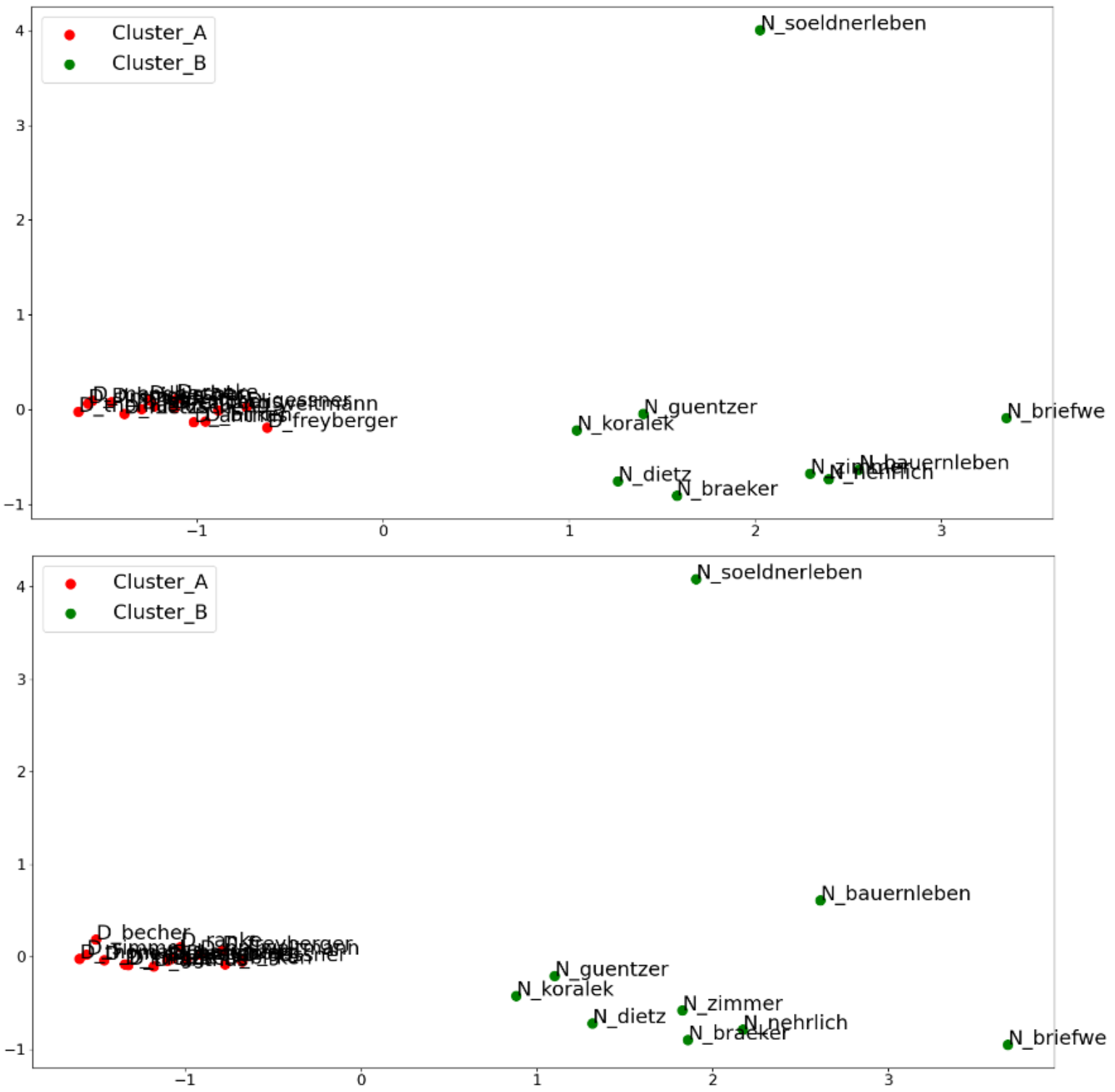

Fig. 8a and 8b: 4 features of conceptual orality – automatic analysis (above) and annotations in the GiesKaNe corpus (below)

In my opinion, this results in the need to distinguish between features of conceptual orality and literacy in terms of which scale range they positively describe. In a first selective approach, I expand the features of conceptual orality to include those of conceptual literacy.

With reference to the external features of the model by Koch and Oesterreicher (1985) before any consideration of statistical multicollinearity, it should be taken into account that they are intended to reflect different conditions of communication. They do not have to represent direct counterparts to the features of conceptual orality. However, such a comparison can be a starting point.

Complexity in the sense of syntactic depth ($x_n$ governs $x_{n+1}$ and $x_{n+1}$ governs $x_{n+2}$ ...) could be contrasted here with linear sentence length in token. In addition to the domain sentence, the noun phrase can be used here. Nominal complexity becomes particularly relevant from New High German onwards (Ágel, 2000; Hennig & Meisner, 2023). Biber and Conrad (2009: 167) describe it as “one of the most dramatic areas of historical change in English over the past three centuries” and Polenz (1999: 353) sees a ‘shift in emphasis from the educational, academic and administrative style (subordinate clauses, maximum hypotactic structures, verbal bracket) to the nominal style (expansion and juxtaposition of noun phrases)’. Nominal complexity is analyzed here via the pre-nominal area (pred-modifiers), so that the automatic analysis can build on a stable segmentation using the head noun as a right bracket, because sequences of post-nominal attributes tend to be ambiguous. All attributes are counted based on their syntactic depth: Each direct attribute of the core noun other than unexpanded adjectives receives the value 1.[15] Each attribute subordinate to these attributes receives the value 2 per occurrence. A participial attribute with a subordinate prepositional attribute receives (1 + 2 =) 3 points (automatic analysis versus annotation: r = 99.2, p < 0.001). Such a view is based on the principles described in Hennig (2020: 151ff.), but also on Biber et al. (1999: 571ff.). These assumptions are also in line with the analysis of Botha and Zyl (2021) on 'Register and modification in the noun phrase', since the pre-nominal participles here correspond to the post-nominal ones there. Botha and Zyl (2021: 204) note:

> "In each of the top-ranking registers pairs […] the register with the higher proportion of modifying -ed/-en clauses would be on the literate-informational end of Biber and Egbert's (2016) Dimension 1, whereas the register with the lower frequency would be oral-involved."

Just as complexity in sentence length can only be measured relative to the predicates, it is crucial here that the creation of scenarios is transferred from the verbal to the nominal domain via participles. Further starting points for the corpus linguistic and/or automatic analysis of syntactic complexity are shown by Hsiao et al. (2024), Châu & Bulté (2023), Diez-Bedmar & Perez-Paredes (2020). By focusing on the noun phrase, the present approach also ensures that complexity describes a dimension other than linear sentence length and information density (see, for example, Lu, 2009, 2010; Osborne, 2011). When further processing the features of conceptual literacy, ung-deverbalization (Hennig & Meisner, 2023)[16] as an indicator of the shift from the verbal to the nominal domain should also be considered – and also nominal compounds (Degaetano-Ortlieb, 2021).

As a counterpart to the 'speaker-listener pronouns' of the first and second person, an impersonal style – 'author-evacuated' (Geertz, 1983), 'deagentivierend' (Hennig & Niemann, 2013) – can be used (see also: Callies, 2013; Harwood, 2005) – here: generic personal pronoun *man*, auxiliary verbs to express passive voice, causative use of *lassen* (to *let*), *sein* (*to be*) as a semi-modal verb like *to have to* (r = 98.7, p < 0.001). Contrasted with exbraciation, parenthesis could indicate planning. This feature has already been discussed because there was a negative correlation with the features of conceptual orality. However, parenthesis cannot be reconstructed from the parser analysis, and a neural network for categorizing spans is not convincing in this task.[17] But planning in a broader sense can possibly (r = 91.6, p < 0.001) be operationalized via the distance (mean token count) between left and right verbal bracket (cf. Schildt, 1976; critically: Ágel, 2000; see also Elspaß, 2010 and frequently cited in Vinckel-Roisin, 2015: Auer, 2000; Zifonun et al., 1997; Engel 1977). With the ratio of autosemantics (without verbs) to the predicates that open up slots for constituents in the first place, a level of information density is taken into account (r = 97.5, p < 0.001), which corresponds to the ratio of full verbs to nouns in Ortmann and Dipper (2019) and Broll and Schneider (2023) and ultimately implements the initially introduced considerations of Biber (2014) (see also Halliday, 1989). Junctors (r = 91.9, p < 0.001) as cohesive means could highlight another dimension of the text (argumentative or narrative, aggregative or integrative, also degrees of planning). This applies to adverbial conjunction and subordinating conjunctions in the mass rather than to conjunctions. But in this approach, no semantic subcategories of junctors are distinguished. In the further analysis (see Fig. 10), junctors as binary predictors clearly lag behind the other values or, purely quantitatively and as a heterogeneous group, not suitable for predicting conceptual orality and literacy. Therefore, they are already excluded from the PCA plot in Fig. 9.

---

[15] As Biber and Conrad (2009: 168) show, this restriction would not be absolutely necessary. By dispensing with simple adjectives, an attempt is made to weight syntactic depth (i.e. values greater than 1) more heavily than large sums of simple adjectives.

[16] Biber (a.o. 1988: 227) Biber basically discusses nominalization with -tion, -ment, -ness or -ity and interprets them to the effect that "they tend to co-occur with passive constructions and prepositions and thus interprets their function as conveying highly abstract (as opposed to situated) information."

[17] I will therefore refrain from taking a look at the discourse. However, parenthesis is increasingly being discussed (by no means wrongly) as a feature of conceptual orality. In contrast, the empirical values show a clear accumulation in conceptual literacy. Therefore, the focus must shift to the polyfunctionality of the structure. It is also questionable whether the parenthesis can be operationalized for the analysis of conceptual orality and literacy, i.e. whether it initially shows differences.

The effect of a one-sided feature selection is also evident in Figures 9, even if it does not immediately stand out at first glance. This is because a strong second principal component distributes the texts along the y-axis, and the color-based cluster assignment occurs more prominently in a borderline area. The second principal component is shaped by the contrast between Simmel and Nietzsche on the one hand and NN-Weltmann, which also persists in the final balanced PCA with 9 features within the cluster of conceptually written texts (see also the values in Fig. 12). However, if one focuses on the name prefixes "D" for German *Distanz* (Engl.: immediacy, conceptually written), these texts are described within a value range of below -3 to approximately 1.5, while the conceptually oral texts cluster within a value range of about 0.5 to 2 (in the upper-right corner).

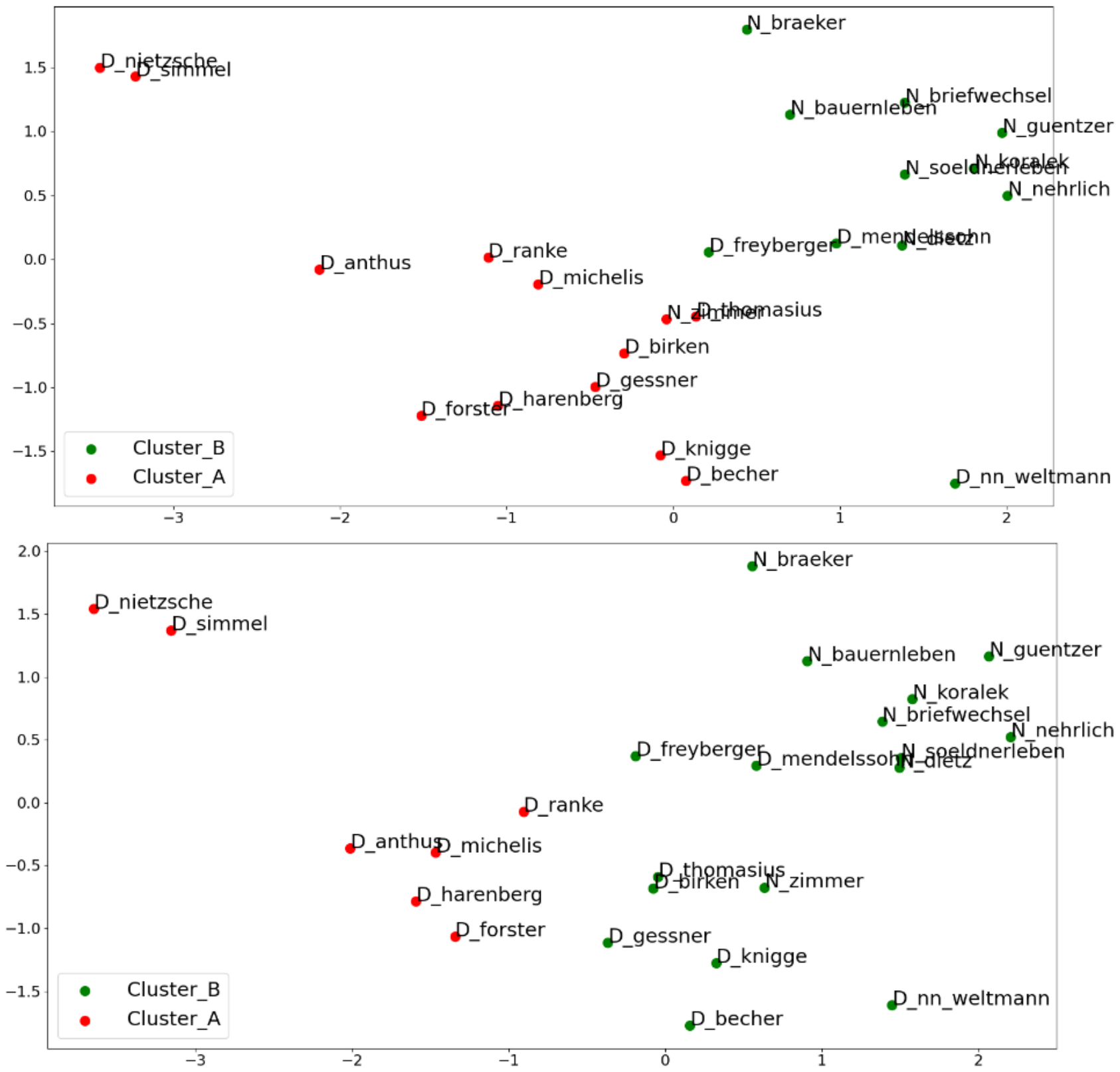

Fig. 9a and 9b: 4 features of conceptual literacy – automatic analysis (above) and annotations in the GiesKaNe corpus (below)

Because in the sense of reversing the analysis steps, binary logistic regression (see Fig. 10) now shows that the sheer count of connectors is too general. A restriction to certain semantic subclasses would involve theoretical definitions that would lead too far away from the actual focus of this article. A starting point for further elaboration is the article by Ágel (2012). Complexity is maintained with weaker $R^2$ and p-value. The implementation of this feature should be revisited in the further theoretically based elaboration of the features of conceptual literacy, although the weaker values in themselves do not represent an exclusion criterion.

| | LR-statistic | df | p ≈ | p-classes | $R^2$ |
|---|---|---|---|---|---|
| syntactic complexity | 6.5 | 1 | 0.01107 | * | 0.32 |
| impersonal style | 11.6 | 1 | 0.00065 | *** | 0.52 |
| information density | 20.2 | 1 | 6.925e-06 | *** | 0.78 |
| finit non-finit distance (verbal bracket) | 15.3 | 1 | 9.026e-05 | *** | 0.64 |
| connectives | 3.4 | 1 | 0.064 | . | 0.18 |

Fig. 10: features of conceptual literacy as binary predictors – tested in the annotations in GiesKaNe

At the end of this section, 9 features are ultimately brought together in PCA and cluster analysis (Fig. 11). The automatic analysis in Fig. 11a shows the position of the texts as they emerge from annotations in GiesKaNe (Fig. 11b). The analysis is completed by a table (Fig. 12), which shows the vectors as a result of the automatic analysis (Fig. 11a) – text vectors as rows and features as columns. The value ranges are highlighted in color for each column/feature, once again clearly showing the difference between the two feature groups. Further analysis of these 9 features shows strong correlation.

The final 9-feature plots (Fig. 11) show a clear separation of the two clusters on the horizontal axis and yet variation: the cluster of conceptually written texts covers an x-axis range of almost 3 units; that of the conceptually oral texts varies by about 2 units. PC1 still represents 52% of the variation in the data. The exact values are provided in Fig. 13. Compared to the representations with only features of conceptual orality or conceptual literacy, with a counterweight variation appears relativized and less extreme. PC2 still explains 17% of the variation and it becomes clear that even with such a tailored selection of features as here, ‘Nähe und Distanz’ only specifies one tendency in one dimension of the text alongside others.

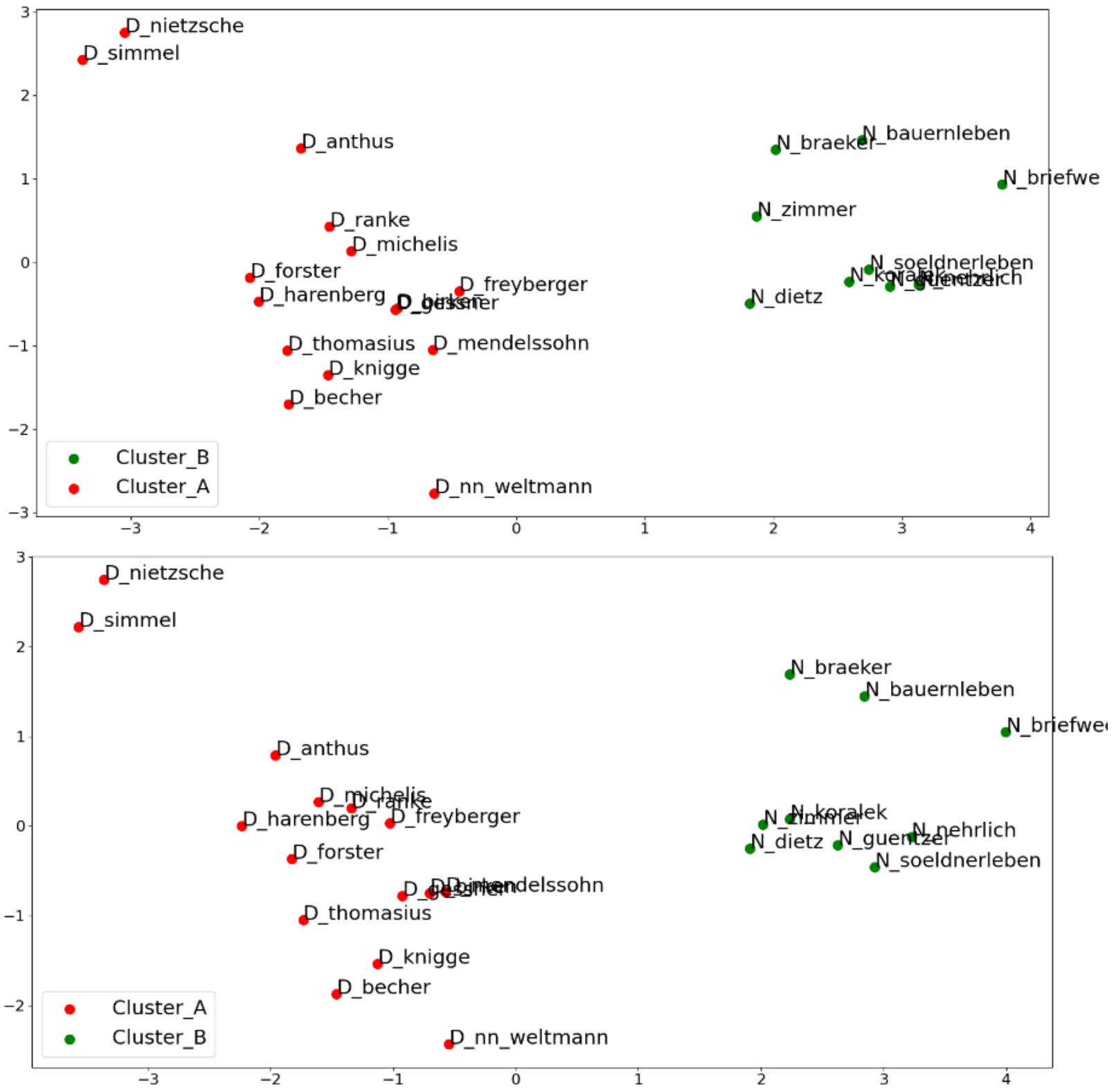

Fig. 11a and 11b: 9 features of “Nähe und Distanz” – automatic analysis (above) and annotations in the GiesKaNe corpus (below)

In comparison with the initial plot of PCA with the 21 features by Ágel and Hennig (2006) (Fig. 3), it is clear in the B/N cluster that the positions of the texts that are far away from the A/D cluster (Briefwechsel, Nehrlich (Hegedüs, 2006), Güntzer, Söldnerleben, Bauernleben) persist with only small position shifts despite the ‘removal’ of many features. Simmel is still the text with the strongest expression of conceptual literacy and this position is now much more pronounced compared to that in Fig. 3 because of the added features with a corresponding focus. But because of this focus, further comparison with Fig. 3 is not very meaningful. With reference to the theoretical model, the 9 features represent operationalizations of the linguistic strategies mentioned by Koch and Oesterreicher (1985) and these are intended to represent external features of communication. These 9 features are the basis for the ranking of written texts on a scale of conceptional orality and literacy.

| | syntactic complexity | impersonal style | information density | finit non-finit distance (verbal bracket) | gram. sentence length | first and second pers. pronouns | temporal adverbials | interjections | exbraciation |
|---|---|---|---|---|---|---|---|---|---|
| D_anthus | 393 | 181.4 | 3.7 | 5.8 | 14.6 | 138 | 143 | 4 | 10.3 |
| D_becher | 156.1 | 229.1 | 3.1 | 5.7 | 22.5 | 59 | 208 | 0 | 8.4 |
| D_birken | 170.8 | 163.7 | 3 | 6.7 | 15.7 | 153 | 221 | 0.8 | 3.2 |
| D_forster | 191.8 | 204.5 | 3.7 | 7.1 | 17.7 | 171 | 174 | 0 | 6.3 |
| D_freyberger | 156.4 | 135.9 | 3.2 | 5.7 | 17.9 | 74 | 313 | 0.9 | 2.6 |
| D_gessner | 208.7 | 194.9 | 3.2 | 6 | 16.4 | 223 | 241 | 0 | 10.6 |
| D_harenberg | 229.4 | 196.1 | 3.2 | 6.8 | 19.2 | 125 | 165 | 0.7 | 5.8 |
| D_knigge | 221.5 | 217.5 | 2.8 | 5.8 | 17.6 | 181 | 126 | 0 | 6.4 |
| D_mendelssohn | 134.3 | 127.9 | 2.9 | 5.2 | 17.9 | 170 | 93 | 0.8 | 8.4 |
| D_michelis | 245.5 | 171.6 | 3.6 | 5.5 | 17.1 | 229 | 153 | 0.8 | 8.7 |
| D_nietzsche | 367.6 | 90.5 | 4.4 | 7.6 | 21.2 | 207 | 118 | 0.8 | 3.3 |
| D_nn-weltmann | 152 | 220.4 | 2.2 | 4.6 | 23.2 | 354 | 166 | 0 | 8.6 |
| D_ranke | 255.1 | 155.9 | 3.6 | 6 | 17.4 | 103 | 216 | 0 | 10.7 |
| D_simmel | 396.7 | 97.3 | 4.2 | 7.2 | 24.3 | 108 | 133 | 0 | 4.2 |
| D_thomasius | 175.6 | 153.1 | 2.9 | 6.1 | 22.7 | 156 | 111 | 0 | 0 |
| N_bauernleben | 158.6 | 91.3 | 3.3 | 4.8 | 13.6 | 345 | 358 | 20 | 60.1 |
| N_braeker | 210.7 | 52.1 | 3.1 | 5.2 | 14 | 687 | 384 | 6.4 | 6.4 |
| N_briefwechsel | 198.4 | 97.1 | 3 | 3.6 | 13.7 | 960 | 410 | 11.8 | 76 |
| N_dietz | 120.7 | 114.1 | 2.5 | 5.5 | 15.2 | 601 | 345 | 6.7 | 10.8 |
| N_guentzer | 110.5 | 79.5 | 2.6 | 4.5 | 11.1 | 870 | 294 | 3.3 | 38.5 |
| N_koralek | 226.5 | 108.7 | 2.3 | 3.8 | 10.2 | 530 | 336 | 5 | 29.3 |
| N_nehrlich | 114.7 | 94.6 | 2.3 | 4.9 | 14.5 | 794 | 364 | 11.2 | 34.5 |
| N_soeldnerleben | 80 | 121 | 3.5 | 4 | 13.6 | 450 | 334 | 1.5 | 226.7 |
| N_zimmer | 241.6 | 173.4 | 3 | 5.3 | 13.3 | 597 | 463 | 9.3 | 30.3 |

Fig. 12: 9 features of 'Nähe und Distanz', data set from the automatic analysis of the 24 GiesKaNe texts, values ≠ quotients are given as relative frequencies (count of hits divided by (count of tokens / 10,000), color highlighting only affects the ratios in one column

In the model by Koch and Oesterreicher (1985: 23) (see Fig. 1), the following scale (Fig. 13) can be equated with the dimension d (printed interview) to k (administrative regulation). Here, however, there are no abstract text types, but rather the concrete texts Briefwechsel, an exchange of letters from the Franco-Prussian War, and Simmel, a scientific text (see Fig. 13). Following the sequence in the theoretical model of Koch and Oesterreicher, the sign of the actual results of the PCA was changed. In the context of this article, it should be taken into account that the values in Fig. 13 can only be compared with the previous 2-dimensional plots with a change of sign: here Briefwechsel is at -3.8 on the left and in Figs. 11 with a value of +3.8 on the right in the plot.

With the present analysis, the findings of Ágel and Hennig (2006) are fundamentally confirmed. Nietzsche, Zimmer, Bauernleben and Briefwechsel exhibit very comparable relationships to one another. However, differences between their absolute and relative values also become apparent. Using 21 of the features of conceptual orality identified by Ágel and Hennig (2006) (see Fig. 3), Briefwechsel, a correspondence (Molnár & Zóka, 2006; 39.3 in Ágel & Hennig, 2006), Söldnerleben, a diary (Peters, 1993; 43.4 in Ágel & Hennig, 2006), and Bauernleben, a chronicle (Mánássy, 2006; 35.3 in Ágel & Hennig, 2006), position themselves uniformly on the scale at a value of 6, distinctly separated from the cluster of conceptually literal texts at approximately -2. Here, Briefwechsel forms the pole of conceptual orality. Güntzer, an autobiography (Kappel, 2006; 38.6 in KAJUK), and Koralek, a diary (39 in Ágel & Hennig, 2006), remain close to each other but position themselves between Briefwechsel, on the one hand, and Söldnerleben and Bauernleben, on the other. Söldnerleben is more moderately positioned alongside Bauernleben. This is well-reflected in the raw base values of the vectors, where Briefwechsel stands out with its dialogic, personal style, characterized by frequent use of pronouns, minimal passive constructions, and temporal adverbials. By contrast, Söldnerleben exhibits extreme values for exbraciation but also shows higher information density, which aligns with its overall profile: Briefwechsel and Söldnerleben display comparable mean sentence lengths and similar distances in the verbal bracket, yet Söldnerleben demonstrates higher information density due to its post-posed information, particularly evident in its enrichment with location descriptions in the context of travel. Bauernleben, with a similarly high information density, features fewer cases of exbraciation but greater distances in the verbal bracket. All three texts, in an initial approximation, are clear examples of conceptual orality. But their raw values already provide a solid foundation for more differentiated analysis.

| | | | |
|---|---|---|---|
| -3.8 | N_briefwechsel | Correspondence | 19 |
| -3.1 | N_nehrlich | Autobiography | 18 |
| -2.9 | N_guentzer | Autobiography | 17 |
| -2.7 | N_soeldnerleben | Diary | 17 |
| -2.7 | N_bauernleben | Chronicle | 17 |
| -2.6 | N_koralek | Diary | 19 |
| -2.0 | N_braeker | Autobiography | 18 |
| -1.9 | N_zimmer | Diary | 19 |
| -1.8 | N_dietz | Autobiography | 18 |
| 0.5 | D_freyberger | (Science) History | 17 |
| 0.6 | D_nn-weltmann | (Use) Decency Lit. | 17 |
| 0.7 | D_mendelssohn | (Science) Theology | 18 |
| 0.9 | D_birken | (Use) Society | 17 |
| 0.9 | D_gessner | (Science) History | 18 |
| 1.3 | D_michelis | (Use) Travel Lit. | 19 |
| 1.5 | D_ranke | (Science) History | 19 |
| 1.5 | D_knigge | (Use) Decency Lit. | 18 |
| 1.7 | D_anthus | (Use) Society | 19 |
| 1.8 | D_becher | (Science) Economics | 17 |
| 1.8 | D_thomasius | (Science) Philosophy | 17 |
| 2.0 | D_harenberg | (Use) Popular Science | 18 |
| 2.1 | D_forster | (Science) Travel Lit. | 18 |
| 3.1 | D_nietzsche | (Science) Philosophy | 19 |
| 3.4 | D_simmel | (Science) Sociology | 19 |

Fig. 13: 1-dimensional ranking of the 24 GiesKaNe texts by automatic analysis (compare Fig. 11a); in the lines: centuries, genre, short title with binary classification, COL value

In comparison to Ortmann and Dipper (2024: 27), who only use first-person pronouns, the implementation of the feature could already have an effect on a different positioning. It is also noteworthy that Zimmer, a diary (Macha, 2001; 29 in Ágel & Hennig, 2006), also clearly sets itself apart from the group around Güntzer, because in my opinion only this text in the KAJUK corpus shows a noticeable deviation from the group around the maximum and, with a value of 29, it moves slightly towards the texts of conceptual literacy around the value zero. Ortmann and Dipper's (2024: 33) classification of Nietzsche and Thomasius – with a slight distance from each other – at the pole of conceptual literacy is comparable while in Ágel and Hennig (2006) the values of Thomasius and Nietzsche are almost identical – with values of 2.6 and 4.1. In Ortmann and Dipper, Zimmer also stands out slightly from the group of texts of conceptual orality. However, there – unlike in Ágel and Hennig (2006) and here – Güntzer achieves the highest value of conceptual orality. Koralek joins the group around Briefwechsel and Bauernleben, which, according to the calculation method of Ortmann and Dipper, are very close to each other.

The further comparison with the value scale from Ágel and Hennig (2006) via Fig. 3 is only partially meaningful because only 21 features were implemented in Fig. 3. Given the different calculation methods, the similar numerical ratios between the text values confirm the features used rather than call them into question – especially against the background of smaller deviations on a large scale. More precise comparisons between COL-value and external features are necessary here. A first attempt is made in section 3.

Since the result of the practical implementation of the theoretical model of 'Nähe und Distanz' must also be understood as a close approximation to Biber's data-driven determination of dimension 1, I will compare both in more detail. With regard to the calculation method itself, Biber's calculation via CFA was reproduced based on the original data set as shared in Biber (1988), whereby the ratios of the genres in dimension 1 correspond exactly to those in Biber (1988: 128): Python's factor analyzer module with Promax Rotation and Maximum Likelihood was used as a fitting method. Applied to the 24 texts of the GiesKaNe corpus, a one-factor CFA shows only minimal, insignificant deviations from the distribution of the texts after applying PCA. As I said, the difference between PCA and CFA is not crucial here and Biber's approach (1988) is not about dimension reduction, but the selection of features and their distribution across dimensions that are thereby defined. Biber (1988) lists 34 features for the first factor. In a further comparison, 33 of the 34 features that form dimension 1 in the genres used by Biber (1988: 102) were reproduced in the GiesKaNe corpus: A feature hedges was omitted alongside similarly queryable discourse particles. Instead of stranded prepositions, an at least approximately comparable structure was chosen with the exbraciation – comparable in terms of the "description of the functions associated with the feature" (Biber, 1988: 221). Biber describes this relation here as "mismatch between surface and underlying representations" (Biber, 1988: 243). The past and present participles WHIZ deletions have been replaced by the corresponding participle attributes.

| | | |
|---|---|---|
| Autobiography | N_nehrlich | 2.20 |
| Correspondence | N_briefwechsel | 2.12 |
| Diary | N_koralek | 1.71 |
| Autobiography | N_braeker | 0.92 |
| (Use) decency Lit. | D_nn_weltmann | 0.64 |
| Chronicle | N_bauernleben | 0.53 |
| Autobiography | N_dietz | 0.44 |
| Diary | N_zimmer | 0.36 |
| (Science) Philosophy | D_thomasius | 0.16 |
| Diary | N_soeldnerleben | 0.15 |
| (Science) Economics | D_becher | 0.08 |
| Autobiography | N_guentzer | 0.01 |
| (Science) History | D_gessner | -0.13 |
| (Use) decency Lit. | D_knigge | -0.15 |
| (Science) Theology | D_mendelssohn | -0.19 |
| (Use) Society | D_birken | -0.50 |
| (Science) History | D_freyberger | -0.52 |
| (Use) Society | D_anthus | -0.77 |
| (Use) travel literature | D_michelis | -0.87 |
| (Science) History | D_ranke | -1.06 |
| (Use) Popular Science | D_harenberg | -1.09 |
| (Science) Travel Lit. | D_forster | -1.18 |
| (Science) Sociology | D_simmel | -1.34 |
| (Science) Philosophy | D_nietzsche | -1.53 |

Fig. 14: CFA on the basis of 34 Dim-1-Features of Biber (1988: 102 ) with 24 GiesKaNe texts
(D and N indicate the categorization of the human raters)

Fig. 14 shows a clear deviation from the distribution in Fig. 13. The ends of the scale are the same. However, the alternation of texts of conceptual orality and literacy (dark grey) calls into question the relationship of Biber's features to the dimension of 'Nähe and Distanz' or the equating of dimension 1 and 'Nähe and Distanz', because the binary categorization of human raters is clearly contradicted and also because the respective genres themselves contradict such a ranking of the texts. It is striking here that NN-Weltmann is positioned squarely among the conceptually oral texts, while Thomasius and Becher exhibit a similar tendency. These texts already formed the counterpart to Nietzsche and Simmel on the second principal component, as shown in Figures 9 and 11. Various aspects need to be taken into account here. Sigley stated that a high number of features favors a distribution across several dimensions and that PC1 could then explain less variation in the data. As shown, a targeted selection of features in PC1 results in it being able to explain 52% of the variation in the primary data. With Biber's 33 features PC1 only explains 0.28 of the variation in the data and PC2 still accounts for 0.2. The effect of the three reconstructed features is opposite because the resulting arrangement of the texts appears even more heterogeneous if they are omitted. The 2D plot of the PCA with the 33 features (Fig. 15) then shows graphically that the two clusters are divided by a diagonal. The proportion of PC2 compared to PC1 is high.

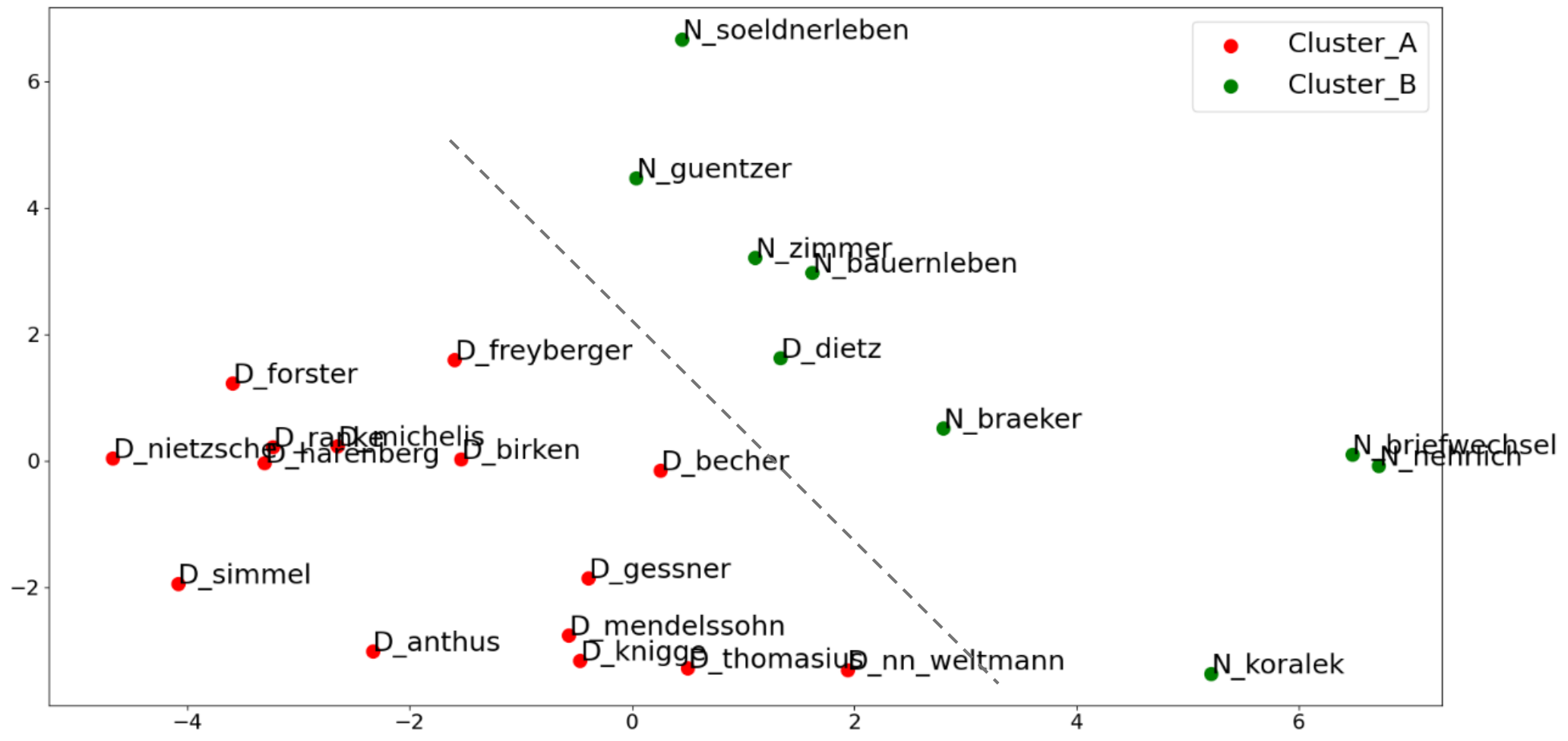

Fig. 15: PCA on the basis of 34 Dimension 1 features of Biber (1988: 102 ) with 24 GiesKaNe texts

This result is not surprising, since the theory-driven model is supposed to be realized as a single dimension of 'Nähe und Distanz', while the data-driven approach captures the texts as what they actually are – a mixture of dimensions – and these in turn represent a simplification of the text's various sub-functions and further conditions of communication.
The relevance of such a quasi-tailored dimension for the analysis arises from the constant application of the model by Koch and Oesterreicher (1985) as its implementation. In practice, the result of the theory-driven starting point, the selection of texts and the selection of the best binary predictors is a 'definition' of PC1 as 'Nähe und Distanz' and in this sense it can be seen as an exaggeration. As mentioned, this results in possible areas of application in supporting research activities such as corpus compilation, orientation in large text collections, and controlling factors surrounding the actual research interest. This comparison also makes this clear. Even if 'Nähe und Distanz' in the present implementation becomes more similar to Biber's approach (1988), different potentials become clear, too.

Compared to the analysis with the prefabricated fast-text vectors in Fig. 2, the cluster of conceptually written texts shows strong parallels that must be addressed as such, but which should not be given too much weight: In Fig. 2, the cluster of conceptually written texts also spans between Simmel and Nietzsche on the one hand and NN-Weltmann on the other, with Becher, Knigge and Thomasius taking a comparable position on NN-Weltmann. The previously mentioned parallels to the analysis with Biber's Dimension 1 features in Fig. 14 are thus also evident here.The position of Freyberger, Gessner and Birken at the 'border' to the cluster of conceptual orality also appears to be well comparable, as does the position of the group around Michelis, Ranke and Harenberg with proximity to Simmel and Nietzsche at the upper edge of the cluster center. Slight deviations from the analysis in Fig. 3 include the less centrally positioned Forster and the more centrally positioned Anthus. The B/N cluster of conceptually oral texts shows some clearer deviations and some similarities: the important positions of Dietz (Rauzs, 2006a) and Briefwechsel are ranked accordingly. In the 9-feature vector model developed here, Briefwechsel has already shown the clearest expression of conceptual orality, which could be attributed to the dialogic structures. The fact that the text stands out even more clearly from the other texts of conceptual literacy when using the fast-text vectors could be justified in an analogous way. However, other aspects seem to be given greater weight in the 300-dimensional fast-text vectors[18], which is not surprising because the vectors were created based on web crawling and Wikipedia articles from contemporary texts. Applied to New High German texts, the conceptually written texts could offer more points of reference here. Ultimately, what I take away from this comparison is that the features of the COL-scale also capture dimensions of the texts that shape them beyond theoretical models. Theoretical ideas about concept and implementation (Koch & Oesterreicher, 1985; Ágel&Hennig, 2006) do not contradict a data-driven analysis of the features of the texts. This means that fast-text vectors are also coming into focus as a linguistic tool. They will be taken up again in section 2.3 when applied to the DTA.

### 2.3 The COL-scale: An Application to the DTA

Even more than comparing it with previous scales and values, the benefit of the COL-scale can be measured in comparison with external conditions of communication. The concept of ‘Nähe und Distanz’ aims to mediate between the internal linguistic features and the external conditions of communication. The extensive DTA corpus, which, as a comprehensive corpus of New High German, fits the scale derived here from the GiesKaNe corpus, offers a relatively free, category-rich form of classification[19] that should actually be more precise for evaluation. But further analyzes will provide enough starting points for further access.

The DTA release ‘TCF-Version vom 19. Oktober 2021, DTA-Kernkorpus und Ergänzungstexte, voll’[20] includes 4.432 documents; 4.209 were taken into account. The missing texts result from the fact that missing values – for example in very short texts (cf. Liimatta, 2023) and/or when eliminating outliers – were not compensated. Here, the idea of evaluating the scale was given greater weight than the analysis of the DTA itself. While binary classification using k-mean cluster analysis is unproblematic (3162 conceptual literacy, 1047 conceptual orality), the application of the scale (mean = 0.78, sd = 1.73) is accompanied by a histogram (Fig. 16a), that raises questions. When ranking the DTA texts according to the scale via PCA, 'new' texts are ranked individually compared to the previous text setup – always 24 + 1. Fig. 16a and also the histogram for the analysis with the fast-text vectors (Fig. 16b) could show a bi-modal distribution. A conceivable alternative would be skewness that would result from the compilation of the DTA corpus. In my opinion, the latter provides an explanation: The compilation of the DTA corpus – described via the length of the grammatical sentence as a central variable here (Fig. 16c) – seems to explain the variance in the data. In contrast, the assumption that the COL-value could be linked to an over dominant feature ‘sentence length’ cannot only be ruled out in linear regression as a reversal of PCA due to significant contributions from all features. The fast-text vectors in Fig. 16b also show the same distribution, but are independent of the modeled features. The comparison with the metadata in Section 3 will show that the 1.300 newspapers, with a mean COL-value of 1.65 (sd=0.68), should be essential to the strength of the second 'peak' shown (Fig. 16a).

---

[18] In considerations of the contrast in the second principal component (Simmel and Nietzsche on the one hand and NN-Weltmann on the other), the contrast between Zimmer and Koralek or Nietzsche should also be included, with additional features beyond the nine in the present model potentially being relevant.

[19] https://www.deutschestextarchiv.de/doku/klassifikation

[20] https://deutschestextarchiv.de/download

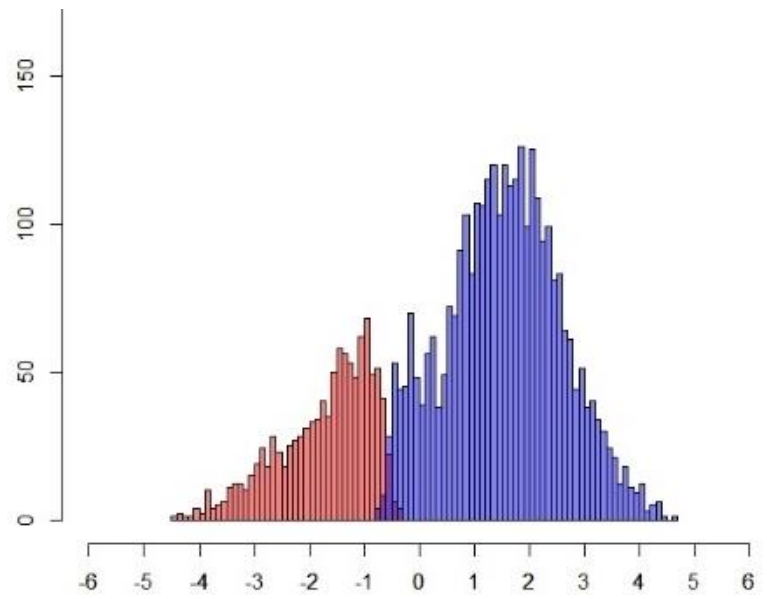

Fig. 16a, 9 features PCA, DTA text count and COL-values, blue=c. literacy, red= c. orality

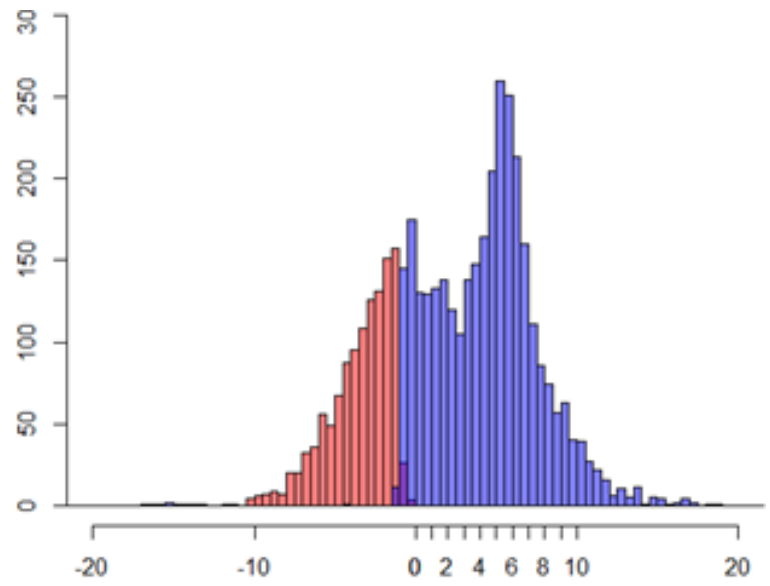

Fig. 16b, fast-text PCA, DTA text count and fast-text vectors, blue=c. literacy, red= c. orality

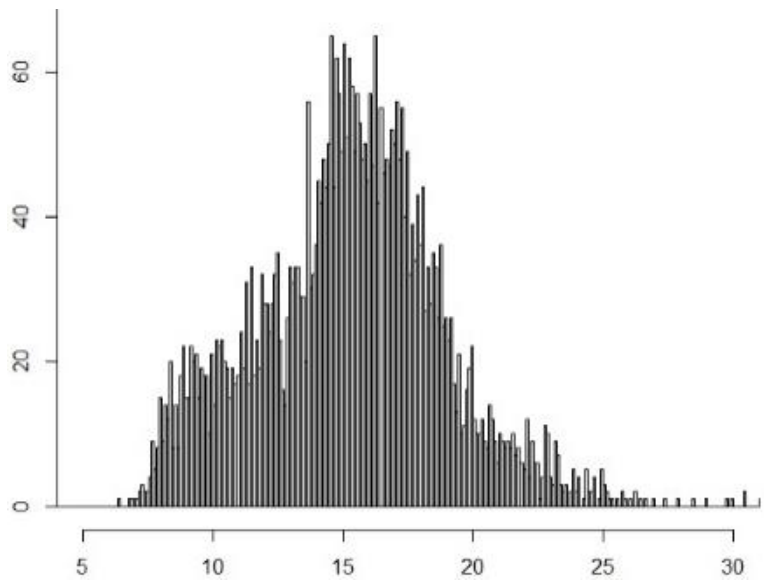

Fig. 16c, length of the grammatical sentence in the DTA

In addition, the sub-topic 'text section vs. entire text' mentioned at the beginning will be treated here. It is elementary for corpus compilation and, as with the BNC (Hundt, 2008: 178), is repeatedly addressed but not empirically analyzed (Sinclair, 1991; Douglas, 2003; Flowerdew, 2004; Sinclair, 2005; Hunston, 2008; Claridge, 2008, Ädel, 2022; Koester, 2022). The problem is illustrated by Biber and Jones (2009: 1289):

> "The number of samples from a text also deserves attention, because the characteristics of a text can vary dramatically internally. A clear example of this is experimental research articles, where the introduction, methods, results, and discussion sections all have different patterns of language use. Thus, sampling that did not include all of these sections would misrepresent the language patterns found in research articles."

Fig. 17 shows the texts of the GiesKaNe corpus contained in the DTA corpus. The total texts were each divided into n segments of around 12.000 token as in the GiesKaNe corpus.[4]

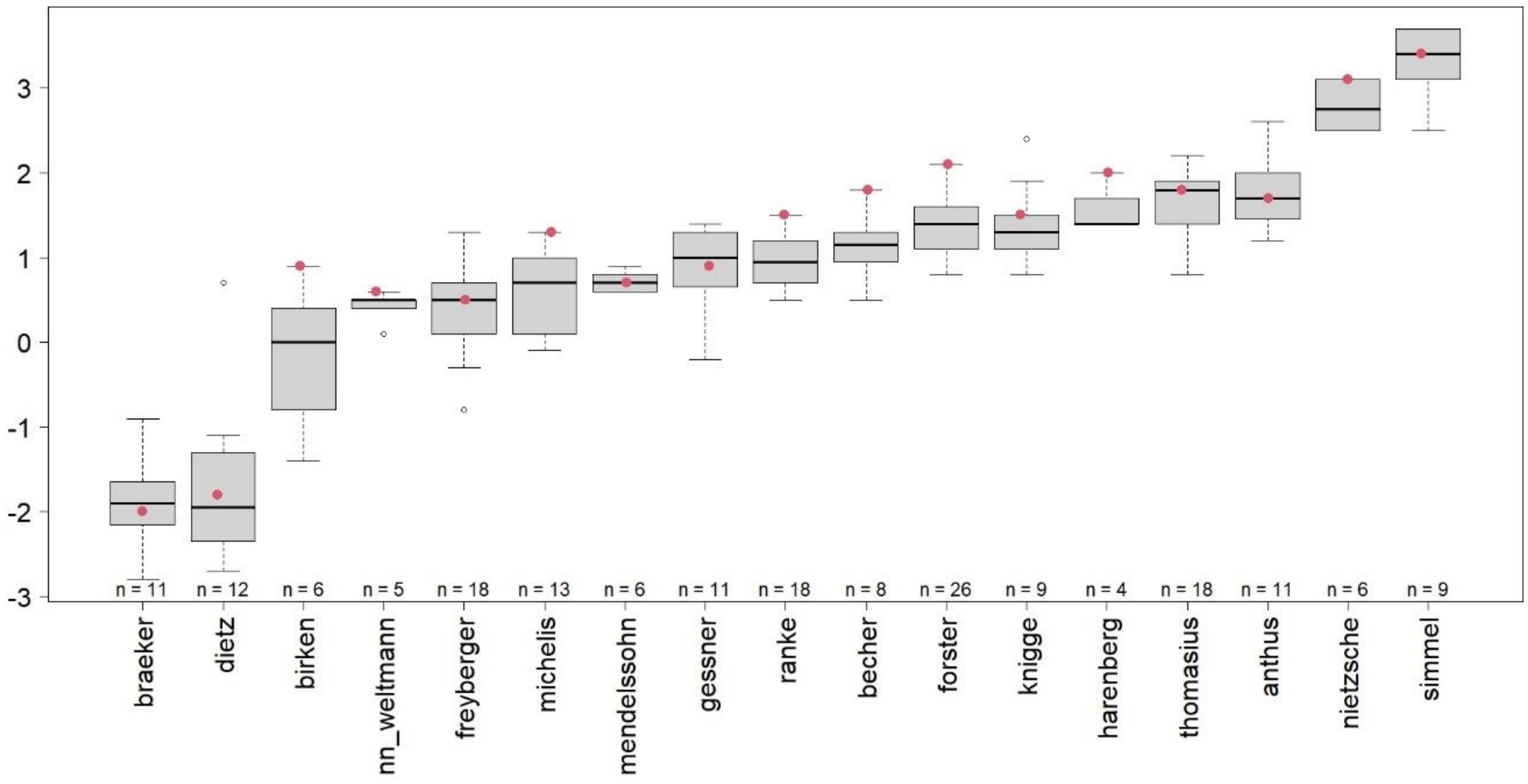

Fig. 17: COL-value of the text segment in GiesKaNe (red dot) in relation to n segments of the same length of the entire text in the DTA corpus

Each segment of a text except the first and the last shares half of the token with an adjacent segment ($Seg_1$ = token 0 to 12.000, $Seg_2$ = token 6.000 to 18.000, $Seg_3$= Token 12.000 to 24.000, ...) to avoid hard breaks. The box plots (Fig. 17) are sorted by the mean, while the black line indicates the median and the red dot indicates the value of the corresponding text segment in GiesKaNe. The mean IQR for the texts is 0.51 (sd = 0.26)and varies between 1.0 for Birken and 0.1 for NN-Weltmann. A higher IQR seems to occur in the texts that tend to be conceptually oral.[21] If the sampling nature of text segments is taken into account, no major problems emerge. Nevertheless, it also becomes clear that special and mostly small corpora should not be viewed without reference to more extensive corpora. As for the cases in which the GiesKaNe text value (red) lies outside the IQR (Birken, Michelis, Ranke, Becher, Forster, Harenberg), this shows that automatic analyzes such as in this approach are essential for corpus compilation – not only in the dimension of 'Nähe und Distanz'. If you look at the IQR or median of the complete texts, the order of the texts on the COL-scale shifts slightly compared to that determined for the GiesKaNe texts (Fig. 13). This does not affect the relationship between features and COL-value, but weakens the possibility of relating feature values and COL-value to the external conditions of communication. As I said: The text segment fully uses the metadata of the entire text, unless further differentiated. This possibility of automatic text analysis in no way reduces the value of qualitative analyzes with a focus on text-functional or thematic aspects, but rather should be seen as a supplement. Sigley (1997: 231) also sees a possible application for his index in the area of micro-analysis of a text, over which a token window of size n is then moved, as here, so that linguistic and functional differences can be identified after analyzing the subsections via the index. This approach is also briefly discussed in Section 3. Additionally/alternatively, the fairly homogeneous value ranges can be seen as an initial indication of the stability of the features in the automatic analysis. But the crucial point here is the comparison of the COL-value with the external conditions of communication.

## 3. An initial evaluation: Internal and external features

The concept of 'Nähe und Distanz' aims to mediate between the internal linguistic features and the external conditions of communication. Further evaluation of the COL-value therefore requires a comparison with these external features. In the research tradition of multi-dimensional analysis, the article by Egbert and Gracheva (2022) can be a good starting point. In 4 case studies they ask how strongly "linguistic variation within registers is influenced by the definition of the textual unit and the situational parameters". Here the extensive DTA corpus, which, as a comprehensive corpus of New High German, fits the scale derived here from the GiesKaNe corpus, offers a relatively free, category-rich form of text classification. Because many (scientific) documents are categorized only by topic, some classes are large (e.g. law, medicine, theology, horticulture). They have been selectively combined here into three groups: humanities, natural sciences, science with a high practical relevance. However, they are not used for evaluation. In practical use, one will probably want to differentiate between the classes of the DTA more precisely.

[21] A visual inspection does not indicate that text length has a noticeable effect on the variation in values. The impact of short texts (cf. Liimatta, 2023) must also be examined in further development.

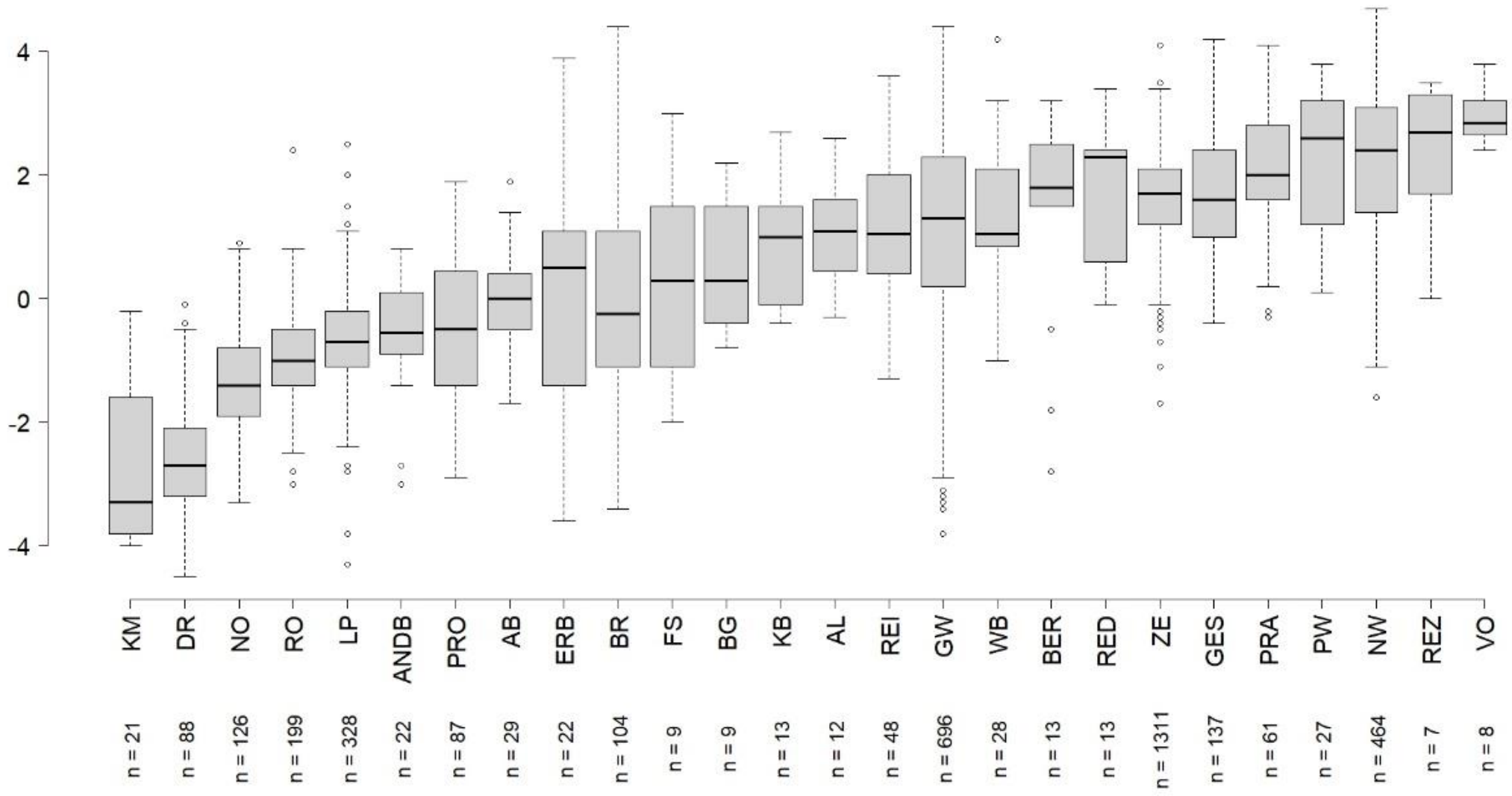

Fig. 18: (simplified) classification of domains and text types in the DTA corpus, sorted by mean of calculated COL-values: **KM** = children's books/fairy tales, **DR** = drama, **NO** = novella, **RO** = novel, **LP** = funeral sermon, **ANDB** = devotional book, **PRO** = prose, **ERB** = edification literature, **AB** = autobiography, **BR** = letters, **FS** = pamphlet, **BG** = biography, **KB** = cookbook, **AL** = moral literature, **REI** = travelogue, **GW** = humanities, **WB** = dictionary, **BER** = report, **RED** = speech, **ZE** = newspaper, **GES** = social writings, **PRA** = practical/science, **PW** = popular science, **NW** = natural science, **REZ** = review **VO** = regulations,

It is clearly visible that the literary genres are associated with very low (conceptually oral) COL-values. This is already shown by the work of Culpeper and Kytö (2000, 2010; cf. Kytö, 2019). Despite the fairly extensive category size (n = 88, 126, 199), the IQR for drama, novella and novel (RO) shows a homogeneous range of values compared to other 'text types' or text classes. This tendency does not need to be explained in detail here, but in principle it represents a complex question about how to deal with imitated orality in texts and corpora. Because there are differences between medial orality and its imitation in fiction (see Jucker, 2021). When applying his formality index to text categories, Sigley (1997: 222 f.) also states that within medial written text categories, literary texts are almost the most informal text types – with a large gap to the other text types. He attributes this result to the imitation of dialogic structures. In principle, the use of such an index also reveals the critical aspect of many corpora: text categories describe very heterogeneous groups of texts. As an alternative, he suggests forming groups of texts or clusters on the basis of linguistic features alone (Sigley, 1997: 227) – as does Gries (2006) (cf. Vetter, 2021). This practical function of the COL-value was also highlighted here. Described as an alternative, in the spirit of 'annotations as added value' it is an alternative of use, not of provision.

Fairy tales (KM) show a larger IQR: here, for example, the fairy tales by the Brothers Grimm (n = 14 texts) show more conceptual orality (-2.8 to -4.0) (mean = -3.5, sd = 0.38) than sagas by Schwab, Temme, and Gottschalck (-0.2 to -1.3). Prose, on the other hand, appears less clearly defined and shows both wider IQR and greater whisker spacing. Funeral sermons (LP) and devotional books (ANDB) show tendencies towards moderately simple language in the features themselves, which could also fit with the external features of these religious texts. But before any comparison, the automatic analysis should be checked for the text type funeral sermon (median year of publication: 1623) and an increased value for the exbraciation. Because Ágel (2000: 1874) notes that exbraciation only occurred to a limited extent in the 17th century. This cannot be done here, but it underlines once again that the COL-scale, like any annotation in corpora, is about an initial orientation with the need for further examination.

Letters (BR) use almost the entire spectrum of the COL-scale, which could be due to the real and therefore varying readers compared to an imaginary reader (on the concept see, among others, Feilke, 2011), whose relationship to the writer is therefore perhaps more stable, such as in academic writing. A letter from Alexander von Humboldt (1854) to Albert Berg, which he uses in his scientific publication, achieves a high COL-value of 3.5, while a letter from Humboldt (1795) to Samuel Thomas Soemmerring (Example 2) has a value of -1,6. This is one of 5 texts by Humboldt rated as conceptually oral in the binary classification – 5 out of 153 with a mean COL-score of 2.44 (sd = 1.16).

> **Example 2:** (letter from Humboldt (1795) to Samuel Thomas Soemmerring)
>
> Ich will Ihnen ein Buch dediciren, ich ein physiologisches; es soll bald gedrukt werden und Sie haben es noch nicht erlaubt..... Das ist eine sonderbare Dreistigkeit. Aber als Ueberraschung wage ich es nicht und die Erlaubniß müssen Sie mir nun schon geben. Könnte ich Sie doch mündlich darum bitten. […]
>
> [engl, not literally:
>
> I want to dedicate a book to you, a physiological one; it should be printed soon and they/you haven't allowed it yet..... This is a strange audacity. But as a surprise, I do not dare do it and you have to give me permission. If I could ask you verbally/in person.]

In general, According to Cvrček et al. (2020), the effect of register is 1.5 times greater than the effect of the author himself. Here, however, it is furthermore about a relationship between writer and reader that is real and private, and the texts, the style of writing or the COL value could change – just like the relationship itself. This can be described by the COL-values of 0.0 and -1.5 in letters from Daniel Sanders to Wilhelm Scherer[22], in which he asked for a report and a discussion of one of his works – same writer, same reader, very similar topic, very similar function, but different time (a few years later), which could mean that the relationship has changed during this time. But the reasons for writing show even greater variation: The previously mentioned letter from Humboldt to Berg (1854) for a scientific publication (COL value = 3.5) versus birthday wishes from Sanders to Adolf Glaßbrenne (1871) with -2.7. Elspaß (2008, 2010), on the other hand, notes when applying the approach of Ágel and Hennig (2006) to letters that no particularly high values of conceptual orality are achieved there, which can be a starting point for a more precise comparison.

In the case of autobiographies, the already mentioned GiesKaNe-texts Bräker and Dietz also have the lowest values of conceptual orality in the DTA, both scoring -1.7 for the entire texts. Bismarck's autobiography ("Gedanken und Erinnerungen", volume 2, 1898) compares with this with a significantly higher COL-value of +1.9: Sentence lengths of 12 and 21 words, verbal-bracket distances of 5 versus 7 words, but in the conceptually oral texts also twice as many pronouns of the first and second person and twice as many temporal adverbials and than no interjections in Bismarck. In an afterword to Schmoller's historical treatment (Schmoller & Krause, 2010: 38f.), Krause supports Schmoller's perception of Bismark's style as remarkably precise and pictorial and cites Schmoller's remark that it is not an artistically rounded representation, but rather a stenographically recorded conversation with Lothar Bucher, whose style Schmoller describes as 'causerie', chat. Schmoller himself (Schmoller & Krause, 2010: 6) adds that these notes were subsequently repeatedly reviewed and changed by Bismark himself. Schmoller sees the text less as a "memoir" and more as a "political textbook," which Krause confirms (Schmoller & Krause, 2010: 40). In this respect, there is also some evidence here that the higher COL-value corresponds to external conditions of communication. Basically, autobiography and biography are close to each other based on their mean values, with biography showing more of a tendency toward conceptual literacy. However, in comparison with the external conditions, the question must be asked how moderate values of conceptual orality can be explained here. Although the sample of n = 9 is too small, there is a slight tendency that there are varieties of the text type in which there are strong tendencies towards conceptual orality: for example in the sense of first and second person pronouns. Eckermann's Goethe biography (1836) ), in several parts, with COL-values of -0.4, -0.6, and -0.8 (twice binary classification as conceptually oral), thus shows a certain ambivalence. It contains, on the one hand, an autobiography of the author himself, and on the other hand, in the spirit of the title "Gespräche mit Goethe" [engl.: Conversations with Goethe], it follows a structure in which the narration is supplemented by direct speech. Here, for example, one finds moderate use of pronouns alongside moderate deagentivization, moderate complexity, and relatively low information density.

[22] Sanders, Daniel (1879): Brief an Wilhelm Scherer. Altstrelitz, 7. November 1879. In: Deutsches Textarchiv. https://www.deutschestextarchiv.de/sanders_scherer_1879/1; Sanders, Daniel (1884): Brief an Wilhelm Scherer. Altstrelitz, 7. November 1884. In: Deutsches Textarchiv. https://www.deutschestextarchiv.de/sanders_scherer_1884/1.

In contrast to the summarized scientific groups, which are difficult to interpret here, newspapers allow new insights. The size of the sample is n = 1311 and yet with an IQR of +0.9 they are one of the most homogeneous text types in the DTA corpus. Since this result also comes from around 14 different newspapers, it could be seen as an indication that conventions established that are connected to the development of a standard. The mean COL-value is 1.64 (sd = 0.68). Additionally, it seems that in the "Neue Rheinische Zeitung" (1848/1849) the supplements can be designed very differently. Content such as a court case (No. 85b) or an election call (No. 198b) form the extremes at the edges of the COL-values for this 'text type'/this medium with values of -1.1 and +4.1. As Hiltunen (2021) shows, further analysis of the variation of sub-registers is promising. Here the 'normal' newspaper editions would be even more centered in terms of the COL-value. This homogeneity is also evident at the feature level in the "Augsburger Allgemeine" (1840): in 181 editions, the average sentence length is between 15.2 and 20.7 (mean = 17.7, sd = 0.9) and $m_{Complexity} = 286.2$, $sd_{Complexity} = 33.1$, $m_{Impersonal} = 165.6$, $sd_{Impersonal} = 15.1$; $m_{Info} = 4.3$, $sd_{Info} = 0.3$; $m_{VB_distance} = 6.5$, $sd_{VB_distance} = 0.6$; $m_{Temporal_adv} = 178$, $sd_{Temporal_adv} = 16.2$; $m_{Exbraciation} = 11.5$, $sd_{Exbraciation} = 3.3$. This in turn speaks for the tendency towards standardization. With this last text type I come to a conclusion about this last evaluation and subsequently about the COL-scale itself.

## 4. Conclusions and outlook

When comparing the COL-value and external conditions of communication, starting points for further analysis could be gained from the variation of the value range within the text types. The further orientation towards the individual features was also used. As far as the automatic analysis in the context of scalar text classification is concerned, the comparison of internal and external features is only a first rough indicator, but the comparison has not revealed any problematic points so far. PCA as a calculation method meets the theoretical assumptions of Koch and Oesterreicher well. This calculation method can also be used to build on the extensive research tradition of multidimensional analysis. As the constant references to multidimensional analysis and register studies make clear, a practical implementation of the model of 'Nähe und Distanz' and the data-driven research on written and spoken language following Biber (1986) are not far apart. However, the different starting points also come with different application potentials. The COL-scale is to be understood as a tool that can be used for corpus compilation, orientation in large corpora and for controlling factors related to the actual research interest. In this respect, this implementation will hopefully introduce a low-threshold point of connection to corpus linguistics from a theory-driven research tradition.

In addition to the evaluation of the scale itself, its application to the DTA revealed (not necessarily new) problems and raised questions: about how to deal with literary texts or only a few conceptually oral texts in the DTA that remain when the literary texts are removed or about how to deal with text categories that are too vague. Through the automatic analysis of the DTA, the smaller, deeply annotated GiesKaNe corpus and the much more extensive, shallowly annotated DTA can be used in a complementary manner. This additional score could make the use of the DTA easier, especially against the background of historical corpora, which are not always accompanied by a balanced, representative compilation. More comprehensive questions where the COL-value could be used include analyzes of diachronic change against the background of simultaneous changes in text types and functional styles.

For the GiesKaNe project, automatic analysis and the COL-scale allow a ranking of the texts, which can supplement the previous binary classification. For (further) corpus compilation, the comparison of external and thematic features can be efficiently supplemented by an analysis of internal features: In order to get an impression of these internal features of the texts before they are included in the corpus, an automatic analysis is essential.

**Sources**